# AURORA-KITTI: Any-Weather Depth Completion and Denoising

Yiting Wang[a,c,d], Tim Brödermann[b], Hamed Haghighi[a], Haonan Zhao[a], Christos Sakaridis[c], Kurt Debattista[a] and Valentina Donzella[d]

[a] *Warwick Manufacturing Group, University of Warwick, Coventry, CV4 7AL, United Kingdom*
[b] *Computer Vision Laboratory, ETH Zürich, Zurich, 8057, Switzerland*
[c] *Photogrammetry and Remote Sensing, ETH Zürich, Zurich, 8093, Switzerland*
[d] *School of Engineering and Materials Science, Queen Mary University of London, London, E1 4NS, United Kingdom*

ABSTRACT

Robust depth completion is fundamental to real-world 3D scene understanding, yet existing RGB–LiDAR fusion methods degrade significantly under adverse weather, where both camera images and LiDAR measurements suffer from weather-induced corruption. In this paper, we introduce **AURORA-KITTI**, the first large-scale, physics-consistent synthetic RGB–LiDAR benchmark for any-weather depth completion and denoising. We further **formulate Depth Completion and Denoising (DCD)** as a unified task that reconstructs clean dense metric depth from weather-corrupted RGB images and sparse LiDAR inputs while suppressing weather-induced noise. AURORA-KITTI contains over *82K* weather-consistent RGB–LiDAR pairs with metric depth supervision, spanning diverse weather types, three severity levels, day/night scenes, paired clean references, lens occlusion conditions, and textual descriptions. To evaluate DCD beyond standard depth-completion accuracy, we introduce LiDAR-aware denoising metrics that measure corrupted-return correction, dropout-region recovery, and reliable-return preservation. Moreover, we provide **DDCD** as a strong and efficient distillation-based reference baseline. DDCD is trained on AURORA-KITTI using a frozen teacher applied to paired clean RGB images only during training; at inference, it uses weather-corrupted RGB and sparse LiDAR. Experiments on AURORA-KITTI and the real-world **DENSE** benchmark show that existing depth completion and foundation depth models remain vulnerable under severe weather, while DDCD improves over weather-aware retraining. Controlled experiments further show that physics-consistent adverse-weather data is the dominant driver of robustness, with clean-prior distillation providing complementary gains. Comparisons with off-the-shelf zero-shot methods use different training protocols; same-data retraining and component ablations separately quantify the contributions of weather-aware data and DDCD. Github: https://github.com/Warwick-Jocelyn/AURORA-KITTI.

## 1. Introduction

Accurate and dense metric depth perception is fundamental for 3D scene understanding across a broad range of real-world applications, such as autonomous driving, robotic navigation, and environmental monitoring [1, 2, 3]. Depth completion (DC) aims to recover a dense depth map from sparse LiDAR measurements, typically assisted by an RGB image that provides rich contextual cues [4, 5, 6, 7]. While modern learning-based approaches achieve strong performance on benchmarks such as KITTI-DC [8], they are largely developed and evaluated under clear-weather assumptions. In real-world deployment, however, sensor suites must operate under rain, snow, fog, and low illumination, where attenuation and backscattering degrade camera images, LiDAR returns, or both. Under such conditions, even state-of-the-art foundation depth estimation models may fail, causing safety concerns (see Fig. 1).

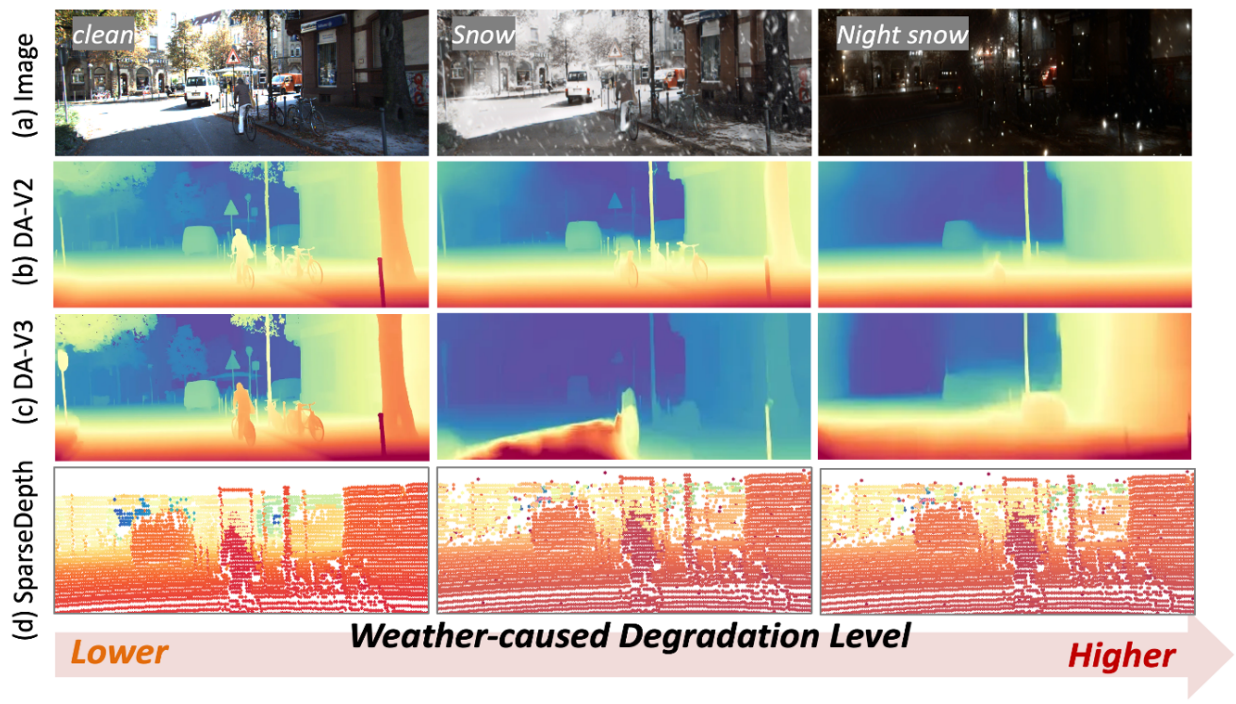

**Figure 1:** Performance of depth foundation models under increasing weather degradation. From left to right: clean, snow, and night-snow conditions. From top to bottom: input images, relative depth prediction (Depth Anything V2), metric depth prediction (Depth Anything V3) and sparse LiDAR map. Severe weather leads to structural distortion and scale instability.

Adverse weather fundamentally alters the sensing process. For cameras, visibility reduction and contrast attenuation obscure structural cues. For LiDAR, weather particles introduce attenuation of valid returns, spurious backscatter-induced false returns, and non-uniform sparsity patterns. Consequently, the problem is no longer pure completion from clean sparse measurements; instead, models must jointly address sparsity and corruption. This observation motivates a task formulation beyond conventional depth completion.

*The work was partially funded by the Centre for Doctoral Training to Advance the Deployment of Future Mobility Technologies (CDT) and the EU Horizon ROADVIEW Project. This work was partially carried out while Yiting Wang was a visiting scholar at PRS, ETH Zurich.

Corresponding author:Yiting.Wang.1@warwick.ac.uk.

ORCID(s): 0000-0002-1268-0168 (Y. Wang); 0009-0009-6428-3066 (T. Brödermann)

**Table 1**
Comparison of existing weather-related RGBL datasets. We summarise weather fidelity across RGB and LiDAR, the availability of compound conditions, paired clean references, raw 3D LiDAR, dense depth labels, the primary task, and the dataset scale. AURORA-KITTI uniquely provides large-scale, weather-consistent RGB–LiDAR pairs with depth labels for DCD.

| Dataset | w-RGB | w-LiDAR | Compound | Paired | r-LiDAR | Depth | Task | Scale |
|---|---|---|---|---|---|---|---|---|
| DENSE-Pixel [9]'19 | ✓ | ✓ | ✓ | × | × | ✓ | depth estimation | 1.6k |
| DELIVER [10]'23 | ✓ | × | × | × | × | ✓ | segmentation | 7.8k |
| WeatherKITTI [11]'24 | ✓ | × | × | ✓ | × | ✓ | detection | 284.3k |
| SemanticKITTI-C [12]'24 | × | ✓ | × | ✓ | ✓ | × | segmentation | 43.5k |
| MUSES [13]'24 | ✓ | ✓ | ✓ | × | ✓ | × | segmentation | 2.5k |
| D-Cityscapes+ [14]'25 | ✓ | × | × | ✓ | × | × | segmentation | 5.0k |
| AURORA-KITTI'26 | ✓ | ✓ | ✓ | ✓ | ✓ | ✓ | depth completion | 82.1k |

We define *Depth Completion and Denoising (DCD)* as the task of recovering a dense, accurate, and clean depth map from corrupted RGB–LiDAR (RGBL) inputs. Unlike standard depth completion, DCD explicitly accounts for multi-modal degradation and targets reliable geometric reconstruction under weather-induced corruption. This formulation is particularly important for real-world sensing scenarios, such as autonomous driving, where consistent metric-scale depth is essential for safe navigation. More generally, it applies to RGBL perception settings in which both modalities are degraded by environmental factors, requiring models to unify depth densification and noise suppression within a single framework.

Despite practical importance, systematic research on DCD remains limited due to data scarcity. Collecting large-scale, dense, and clean depth ground-truth data under real adverse weather is extremely challenging: active sensors suffer from atmospheric scattering and attenuation, stereo-based supervision becomes unreliable under low visibility, and manual annotation of dense metric depth is infeasible. Existing real-world datasets only partially bridge this gap. As summarised in Tab. 1, most datasets satisfy only a subset of the desired properties: some provide weather fidelity in RGB but not LiDAR, others include LiDAR perturbations without paired clean references, and many lack metric depth annotations suitable for depth completion evaluation. Moreover, real-world adverse-weather datasets are typically limited in scale (e.g., DENSE and MUSES have no more than 2.5K samples), scene diversity (e.g., DENSE lacks snowy conditions), or controlled clean references for robustness analysis. This fragmentation hinders systematic training and controlled benchmarking of robust DCD models. Consequently, large-scale training and evaluation under diverse adverse conditions remain largely unexplored.

To address this gap, we introduce **AURORA-KITTI**, the first large-scale, physics-consistent, multi-weather RGBL benchmark designed specifically for DCD. AURORA-KITTI contains over 82K weather-consistent RGB–LiDAR pairs with metric depth labels, spanning multiple weather types, three severity levels, and both daytime and nighttime settings. Unlike datasets that only vary LiDAR sparsity, we explicitly model cross-modally aligned weather degradation across RGB and LiDAR. Using a particle-based rendering process with shared weather parameters (e.g., rain rate and fog attenuation coefficient), we generate weather-corrupted RGB images and sparse depth observations with matched weather type and severity. Specifically, we inject physics-inspired perturbations directly into raw 3D point clouds, simulating attenuation, scattering, and particle-induced backscatter before projecting corrupted points onto the image plane to obtain paired 2D sparse depth maps; see Fig. 3. This process produces realistic non-uniform sparsity, range-dependent degradation, and spurious returns, ensuring consistent severity levels between RGB and LiDAR corruption. We also design an observation-driven lens-occlusion model based on real-world weather imagery to simulate adherent raindrops and snowflakes on the camera lens, introducing spatially localised blur and, for snow, chromatic tint near image boundaries. Each sample further provides a paired clean reference and textual weather description, enabling controlled supervision and robustness analysis.

A natural question is whether recent depth foundation models can already solve DCD. Benefiting from large-scale pretraining, monocular depth foundation models encode strong geometric priors and demonstrate impressive generalisation under clean conditions. However, as illustrated in Fig. 1, their performance degrades increasingly as weather severity increases. While relative depth predictions retain coarse structure, metric depth predictions exhibit noticeable scale instability and structural distortion under heavy snow and low illumination. By contrast, LiDAR projections remain comparatively robust to illumination changes but suffer from sparsity, range attenuation, and particle-induced backscatter. These observations reveal a critical gap: *foundation models provide dense structural cues but are sensitive to environmental conditions, whereas LiDAR offers weather-resilient geometry that is sparse and noisy*. Simply scaling foundation models or naïvely fusing modalities is insufficient. This gap motivates a principled framework that jointly exploits dense structural priors while explicitly modelling completion and denoising under adverse weather.

Leveraging the unique advantages of the AURORA-KITTI dataset, including physics-consistent multi-modal degradation, matched clean references, and metric depth labels, we organise our methodology around two complementary objectives. First, to separate the effect of training data from architectural design, we retrain representative depth models on AURORA-KITTI under the DCD objective without modifying their architectures, isolating data-domain gains from model capacity.

Second, we provide DDCD as a strong and efficient reference baseline for the proposed benchmark. DDCD transfers clean structural priors from a frozen depth foundation teacher using a multi-scale scale-and-shift-invariant matching objective. Its adaptation to DCD is enabled by AURORA-KITTI in two aspects: a GT-aligned teacher normalisation strategy that places heterogeneous relative- and metric-depth teachers into a common metric-depth space, and the use of the paired clean reference image as the teacher input, so that the distilled prior captures weather-free structural geometry rather than degradation-biased predictions. This design offers a practical reference framework for jointly studying completion, denoising, and clean-prior transfer under adverse weather. The clean RGB reference is privileged supervision available to the teacher only during training; the DDCD student uses degraded RGB and sparse LiDAR at inference. Therefore, the clean reference contributes to training-time performance even though it is unavailable at inference. Our ablation studies further show that weather-aware training data provides the dominant robustness gain, while DDCD contributes complementary improvements in structural consistency and metric alignment. These findings highlight AURORA-KITTI as the central enabler for robust DCD research. We distinguish comparisons with off-the-shelf zero-shot methods from controlled comparisons that quantify the gains from AURORA-KITTI retraining and the additional contribution of DDCD under the same training data. Overall, our main contributions are as follows:

- We formulate **Depth Completion and Denoising (DCD)**, a unified task that reconstructs dense, clean metric depth from weather-corrupted RGB images and sparse LiDAR inputs while suppressing weather-induced noise.
- We introduce **AURORA-KITTI**, the first large-scale, physics-consistent, synthetic multi-weather RGB–LiDAR benchmark for DCD, containing 82K+ samples with cross-modally consistent degradation, paired clean references, metric depth supervision, diverse severity and illumination levels, lens-occlusion conditions, and textual descriptions.
- We establish a unified DCD benchmark covering ***10*** representative LiDAR-only, RGB-only, and RGB–LiDAR methods, together with LiDAR-aware denoising metrics. We also provide DDCD as an efficient in-domain reference baseline using paired clean RGB for teacher supervision only during training and show that weather-aware retraining yields the largest robustness gain, with clean-prior distillation providing complementary improvements.

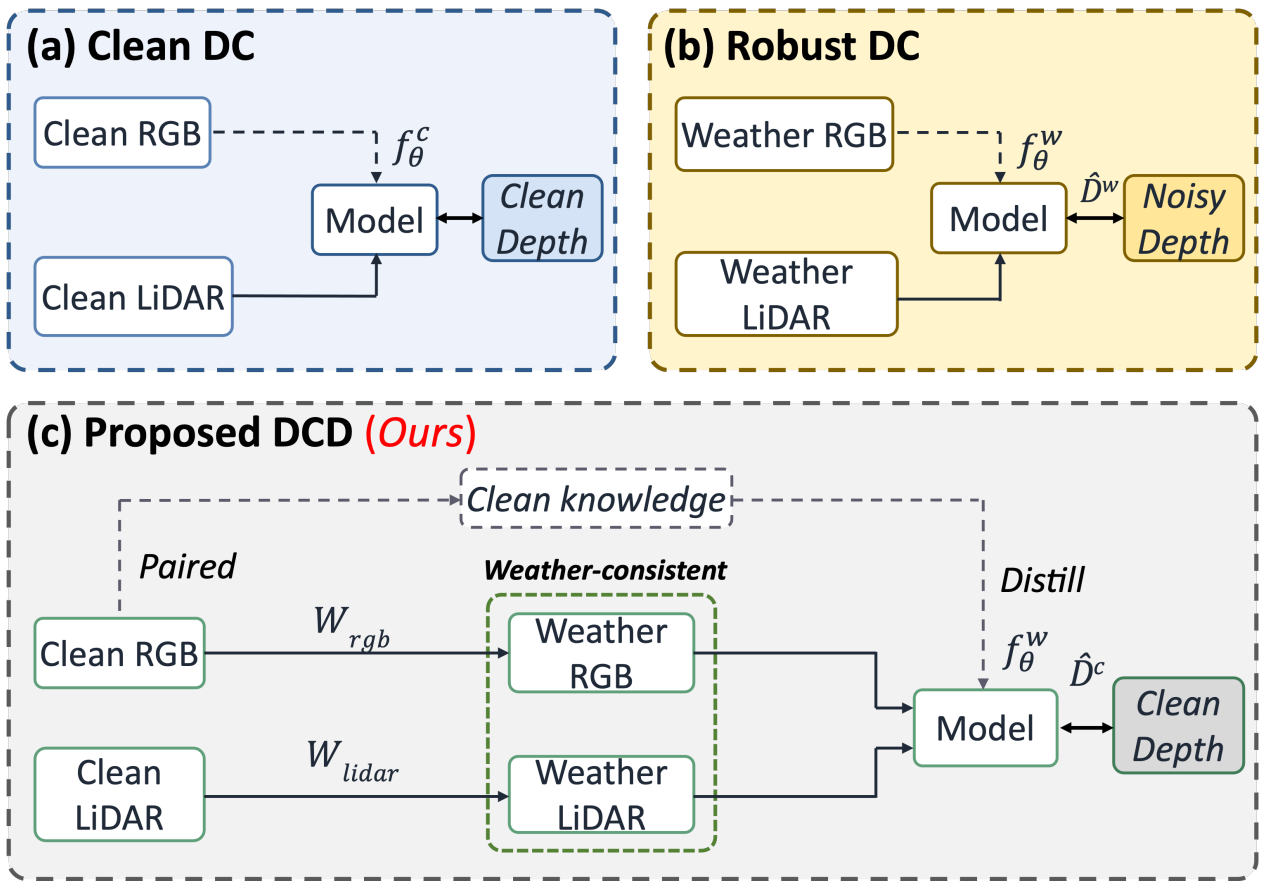

**Figure 2:** The proposed depth completion and denoising paradigm in (c) versus normal-weather depth completion in (a) and robust depth completion in (b).

## 2. Related Work

### 2.1. Robust Benchmarks under Adverse Weather

KITTI Depth Completion (KITTI-DC) [8] remains the most widely used benchmark for depth completion in outdoor environments, providing large-scale RGB–LiDAR pairs with dense labels. However, it is collected primarily under clear-weather conditions. As a result, models trained and evaluated on KITTI-DC often suffer substantial performance degradation when deployed under adverse weather, illumination shifts, or sensor perturbations.

Beyond clean weather depth estimation and depth completion, several benchmarks also investigate the robustness with degraded RGB, degraded LiDAR, or degraded multi-sensor settings. For example, SemanticKITTI-C [12] and D-Cityscapes+ [14] inject adverse-weather effects into LiDAR point clouds or RGB images for robustness evaluation, primarily targeting semantic segmentation and panoptic segmentation. While valuable, these benchmarks do not provide dense depth labels suitable for depth completion. DELIVER [10] includes diverse weather conditions and dense depth maps, but degradation is applied mainly to RGB without modelling weather-consistent LiDAR corruption. MUSES [13] captures real-world multi-modal adverse-weather data across day and night; however, it is designed for segmentation rather than depth completion, lacks reliable dense depth labels, and remains limited in scale (2.5K pairs). Among existing datasets, it is not possible to leverage paired clean sensor data to support denoising in real-world scenes with dynamic and constantly changing objects, nor is it possible to directly obtain clean and dense depth labels under adverse weather conditions. In contrast, the hereby proposed AURORA-KITTI provides large-scale, paired RGB–LiDAR degradations with dense depth supervision, clean references, and weather-consistent multi-modal perturbations, enabling systematic benchmarking for Depth Completion & Denoising (DCD).

## 2.2. Depth Completion and Robust Modelling

Depth completion aims to predict dense depth from sparse LiDAR guided by RGB imagery. Early methods focus on encoder–decoder fusion and sparsity-aware propagation, including Sparse2Dense [5], PENet [15], CSPN++ [16], and NLSPN [17]. Among these, spatial-propagation networks such as CSPN++ and NLSPN iteratively refine an initial dense estimate through learned local affinities and remain strong, widely adopted completion baselines. Recent approaches incorporate global context via attention and transformers, e.g., CompletionFormer [18], or leverage geometric reasoning in 3D space, e.g., BP-Net [7] and TPVD [19]. CompletionFormer couples convolutional locality with transformer-based global context; together, these lines of work span the two dominant design axes of RGB-guided completion: local propagation and global attention. OMNI-DC [20] further improves zero-shot depth completion by processing depth gradients at multi-resolution scales. We therefore include representative propagation-, transformer-, geometry-, and robustness-oriented completion methods in our benchmark.

Parallel to completion-specific designs, monocular depth foundation models have demonstrated strong cross-domain generalisation by leveraging large-scale pretraining and teacher–student learning paradigms, such as MiDaS [21], UniDepth [22], and the Depth Anything series [23, 24, 25]. These models encode powerful dense geometric priors and have been increasingly adopted as guidance signals for downstream tasks such as depth completion and multi-modal fusion [26]. However, they are mainly designed and trained on large-scale datasets captured under clean-weather conditions, showing limited robustness in predicting accurate metric depth under extreme weather corruptions; see Fig. 3. This clean-input assumption also appears in the broader camera–LiDAR fusion literature. For example, BEVFusion [27] fuses camera and LiDAR features in a shared bird's-eye-view representation for 3D perception, while the survey by Cui et al. [28] provides a broader overview of multi-modal fusion. Closer to our setting, Liu et al. [29] study cross-modal 360-degree depth completion for indoor scenes, and Liu et al. [30] extend panorama–LiDAR fusion to outdoor omnidirectional depth. These works demonstrate the benefit of RGB–LiDAR complementarity, but they do not address the setting where both modalities are physically degraded by adverse weather.

Recent efforts have started to explore robustness under degradation by explicitly modelling adverse weather or leveraging auxiliary priors. For example, AWDepth [31] improves monocular robustness via masked-encoding objectives, while SigNet [32] and related works treat degradation as a structured variable in completion. Diffusion-based priors such as Marigold-DC [33] reformulate completion as conditional generation and show promising zero-shot transfer ability; however, the Stable Diffusion denoising mechanism requires long processing time, which is difficult for real-world applications where real-time processing is important. Unlike prior work that either assumes clean sparse inputs or addresses degradation implicitly, we explicitly formulate Depth Completion & Denoising (DCD), where both RGB and sparse LiDAR can be weather-corrupted, and focus on systematic benchmarking with weather-consistent modelling across modalities, using paired clean references only as training-time guidance.

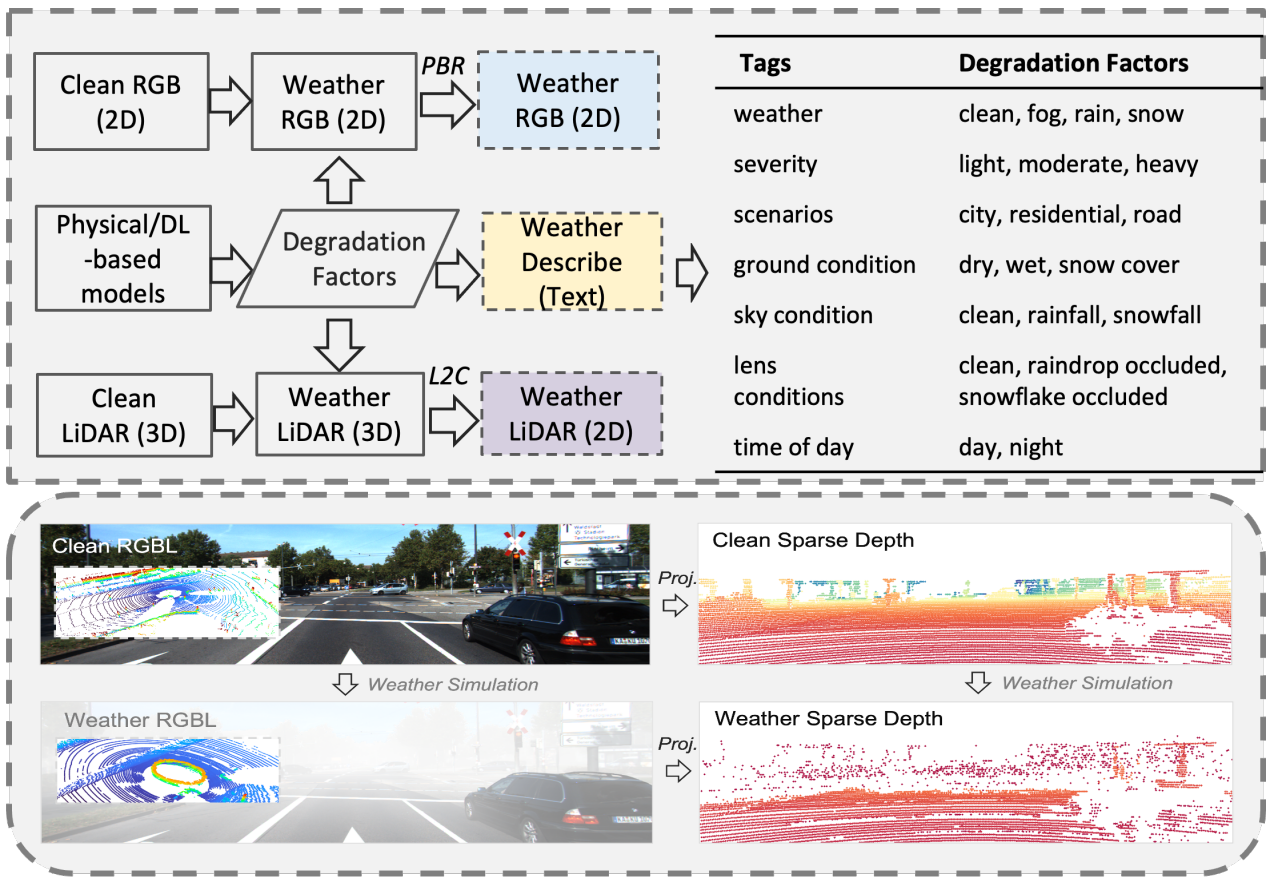

**Figure 3:** The simulation pipeline for both the RGBL and the text prompt information in the AURORA-KITTI dataset. *DL refers to the deep-learning-based methods.*

## 3. AURORA-KITTI Benchmark Construction

This section describes the construction of AURORA-KITTI as a benchmark for Depth Completion and Denoising (DCD). It focuses on the DCD task definition, physics-consistent RGB–LiDAR weather generation, dataset composition, and evaluation protocol. The proposed DDCD model is used as a reference baseline to demonstrate the utility of the benchmark and is described compactly in Sec. 3.5, with implementation-level details provided in Sec. 4.

### 3.1. DCD Task Definition

*Conventional DC.* Conventional depth completion assumes clean multi-modal inputs and clean dense supervision. Given a clean RGB image $I^c \in \mathbb{R}^{H\times W\times 3}$ and its sparse depth from the LiDAR projection $S^c \in \mathbb{R}^{H\times W}$, a completion model $f^c_\theta$ predicts a dense depth map following the equation,

$$\hat{D}^c = f^c_\theta(I^c, S^c), \tag{1}$$

typically supervised by the clean dense ground-truth data $D^c$.

*Weather-corrupted inputs.* Robust depth completion often studies robustness to sparse-depth sparsity, noise, or sampling-pattern changes, but these perturbations are usually not physically coupled with RGB degradation. In DCD, both modalities are instead corrupted by a shared weather state. Let $\mathcal{W}_{\text{rgb}}(\cdot;\omega)$ and $\mathcal{W}_{\text{lidar}}(\cdot;\omega)$ denote weather perturbations parameterised by $\omega$ (type and severity). The weather-corrupted RGB and LiDAR inputs are then formulated as

$$I^w = \mathcal{W}_{\text{rgb}}(I^c;\omega), \qquad S^w = \mathcal{W}_{\text{lidar}}(S^c;\omega). \tag{2}$$

Naïvely applying standard completion on $(I^w, S^w)$ produces *noisy* dense predictions that inherit weather artifacts, degrading boundary fidelity and geometric consistency.

*DCD.* We define the *Depth Completion and Denoising (DCD)* model $f_\theta^w$ as recovering a *clean* dense depth map from corrupted multi-modal inputs with clean depth labels:

$$\hat{D}^c = f_\theta^w(I^w, S^w), \tag{3}$$

where the target $D^c$ corresponds to the same scene under clean sensing conditions. DCD differs from standard completion by explicitly requiring suppression of weather-induced corruption while densifying sparse measurements.

*Difference from existing depth tasks.* DCD is a task-level formulation rather than a generic robustness setting. As summarised in Fig. 2, it differs from existing depth tasks in input degradation, target definition, evaluation metrics, and application objective. Conventional DC assumes clean RGB–LiDAR inputs and focuses on densifying reliable sparse depth. Robust DC mainly studies robustness to sparse-depth sparsity, noise, or sampling-pattern changes, while the RGB input is usually clean or unchanged, and the corruptions are not physically coupled across modalities [17, 16, 32, 20]. Adverse-weather depth estimation instead considers weather-corrupted RGB, but is typically monocular/RGB-centric and does not explicitly model sparse-LiDAR artefacts such as attenuation, dropout, backscatter, spurious returns, and range perturbation [34, 11, 31]. In contrast, DCD corrupts both RGB and sparse LiDAR under the same weather state $\omega$, and supervises the model to recover the clean, dense metric depth from corrupted inputs. Therefore, its evaluation penalises both incomplete geometry and residual weather-induced noise against the clean target. Beyond standard depth-completion metrics, DCD further requires LiDAR-aware denoising evaluation: corrupted-return correction, dropout-region recovery, and reliable-return preservation are measured through ERR, DCE, and CPP, respectively. These metrics explicitly evaluate whether a model can suppress weather-induced sparse-LiDAR artefacts rather than merely densify a noisy input. This makes DCD a joint depth completion and denoising task under physically matched RGB–LiDAR degradation, which is not covered by existing robust DC or adverse-weather depth-estimation formulations.

*Synthetic Supervision.* Acquiring tuples $(I^w, S^w, D^c)$ in real adverse weather is practically infeasible: dense depth captured under adverse conditions is itself corrupted, and obtaining pixel-aligned clean depth for the same scene is unrealistic. This supervision mismatch prevents scalable training and controlled benchmarking for DCD. AURORA-KITTI addresses this gap as a controllable synthetic proxy that preserves clean metric depth supervision and paired clean references, enabling both systematic DCD evaluation and training-only clean-prior distillation.

*Overall Methodology.* Fig. 2 summarises our approach, consisting of three components. **(1) Physics-consistent RGBL dataset generation (Sec. 3.2).** We synthesise weather-consistent RGB–LiDAR corruptions with physically meaningful severity controls, while preserving clean dense depth supervision and paired clean references. **(2) Data-driven robust training (Sec. 3.4).** We use AURORA-KITTI as *in-domain adverse training data* to retrain representative depth completion baselines under the DCD objective, isolating the impact of data coverage from architectural changes and enabling systematic robustness improvement across representative models. **(3) Clean structural prior distillation (Sec. 3.5).** To recover fine structures under severe corruption, we distill dense geometric priors from a depth foundation model teacher (predicted on paired clean RGB) into the DCD network using a scale-and-shift-invariant objective, improving structural fidelity with minimal computational overhead.

## 3.2. Physics-consistent RGB–LiDAR Weather Generation

AURORA-KITTI is designed to support physics-consistent benchmarking and training for depth completion and denoising. In this dataset, both RGB and LiDAR observations are degraded under matched adverse-weather conditions, while paired clean references and dense supervision are preserved. We denote by $\mathcal{P}$ the 3D LiDAR point cloud and by $S$ its projection onto the camera image plane, which forms the sparse LiDAR depth map. Given a clean KITTI-DC sample $(I^c, \mathcal{P}^c, D^c)$, consisting of a clean RGB image $I^c$, a clean 3D LiDAR point cloud $\mathcal{P}^c$, and a dense ground-truth depth map $D^c$, we generate a weather-corrupted tuple:

$$I^w = \mathcal{W}_{\text{rgb}}(I^c; \omega), \quad \mathcal{P}^w = \mathcal{S}_{\text{lidar}}(\mathcal{P}^c; \omega), \quad S^w = \Pi(\mathcal{P}^w), \tag{4}$$

where $\omega$ encodes weather type, severity, time of day, and lens condition. Specifically, $\mathcal{P}^w$ is the weather-corrupted 3D LiDAR point cloud, and $S^w$ is the corresponding sparse depth map obtained by projecting $\mathcal{P}^w$ into the perspective camera view through $\Pi$. $\Pi(\cdot)$ denotes LiDAR-to-image projection using KITTI calibration to obtain the sparse depth map aligned to $I^w$. Crucially, we apply *the same* $\omega$ to both modalities to guarantee cross-modal weather consistency. Table 2 and Fig. 3 summarise the implementations and physically meaningful severity levels.

***RGB-text pair generation.*** We model adverse-weather RGB observations using a layered image formation framework that explicitly separates (i) surface and illumination transformation, (ii) volumetric particle rendering, and (iii) sensor-level lens occlusion modelling. Given a clean image $I^c$, we first simulate global surface and illumination changes (e.g., wet road, snow-covered road, nighttime appearance) using pretrained image-to-image translation models [14, 11].

**Table 2**
**Weather-consistent RGB–LiDAR generation in AURORA-KITTI.** For each weather type, we apply matched severity parameters across RGB and LiDAR using physically interpretable settings (e.g., rain/snow rate, fog attenuation).

| Weather | Sensor | Implementation | Severity / Configs |
|---|---|---|---|
| Nighttime | RGB | CycleGAN-based image translation [14] | Pretrained model from [14] |
| Wet road | RGB | CycleGAN-based image translation [11] | Pretrained model from [11] |
| Rain | RGB | PBR rain rendering with photometric consistency on wet road [35] | Rain rate (mm/hr) $\{10, 100, 200\}$ |
| Rain | LiDAR | Physics-based rain simulation on 3D point clouds [36] (+ wet-road effect [37]) | Rain rate (mm/hr) $\{10, 100, 200\}$ |
| Raindrop (lens) | RGB | Mask compositing (randomised placement/blur/opacity) | Curated droplet masks (artist-designed) |
| Fog | RGB | Atmospheric scattering model [14] | Attenuation coefficient ($\mathrm{m}^{-1}$) $\{0.01, 0.1, 0.2\}$ |
| Fog | LiDAR | Signal-domain fog model on 3D point clouds [38] | Attenuation coefficient ($\mathrm{m}^{-1}$) $\{0.01, 0.1, 0.2\}$ |
| Snow | RGB | Image translation + particle rendering [14, 11] | Snow road + particle masks (3 types) |
| Snow | LiDAR | Physics-based snowfall simulation on 3D point clouds [37] | Snow rate (mm/hr) $\{0.5, 1.5, 2.5\}$ |
| Snowflake (lens) | RGB | mask compositing (randomised placement/blur/opacity) | Curated snowflake masks |

These transformations preserve scene geometry while modifying scene-level appearance shifts. Secondly, we model in-air particle effects. For fog synthesis, we adopt the standard atmospheric scattering model based on Koschmieder's law, which combines scene attenuation and additive airlight under a homogeneous medium assumption [39]. In our implementation, fog severity is controlled by the attenuation coefficient, while the transmittance is determined by the propagation distance from the camera to the scene point along the viewing ray. For rain and snow, we incorporate a physics-based particle rendering engine [35, 14] to generate streaks and falling particles with severity controlled by physically meaningful parameters (e.g., rain rate).

Beyond in-air effects, we explicitly model lens-attached occlusions, including adherent raindrops and snowflakes as observed in real-world sensing deployment. Unlike volumetric particles, these artifacts occur on the imaging system itself and introduce spatially localised, defocused occlusions. To mimic these effects, we design an observation-driven, image-based simulation pipeline that generates such occlusions via mask compositing with randomised spatial placement, blur (to simulate defocus), opacity modulation, and severity-aware layering. For snow, we additionally allow spatially varying chromatic tint to reflect partially opaque and soiled snow accumulation observed in real driving scenes. The third and fifth columns from Fig. 4(a) show that this layered design reproduces characteristic visibility attenuation, streak patterns, and localised occlusions observed in real captures. Full formulation and implementation details are provided in the *Appendix A*. To enable controlled analysis and multimodal research, we provide a structured text annotation for each frame (Fig. 3), encoding weather type, severity, time of day, scene context, and lens conditions. These annotations support attribute-aware evaluation and can facilitate multi-modal conditioned learning under diverse weather conditions.

***LiDAR generation.*** We simulate adverse-weather LiDAR directly in the raw 3D point cloud using established physics-based models with interpretable severity parameters. Rain simulation [36] captures signal attenuation, droplet-induced backscatter, range and intensity perturbations, leading to both point dropouts and spurious returns under heavy rain. Fog simulation [38] models the received response as a combination of attenuated hard-target returns and volumetric backscatter, producing characteristic fog clutter and reduced effective sensing range. Snow simulation [37] explicitly accounts for particle-beam interactions and occlusions, inducing non-uniform sparsity and clutter. For each weather type, we discretise three severity levels using physical parameters (Table 2), and apply the LiDAR projection into the camera field-of-view to ensure cross-modal consistency (Fig. 4). To obtain an RGB-aligned sparse depth map $S^w$, we project $\mathcal{P}^w$ onto the rectified camera frame using the provided KITTI camera intrinsic and extrinsic parameters and rasterise points with a nearest-depth z-buffer rule. The full derivations and implementation details of the 3D-to-2D projection pipeline, including FoV filtering, rectification, and rasterisation, are provided in *Appendix B*.

### 3.3. Dataset Composition and Splits

***Dataset Statistics*** AURORA-KITTI is built on top of KITTI-DC [8] and contains 82,177 RGB–LiDAR pairs with aligned metric depth supervision that is denser than the original sparse LiDAR. Each sample is annotated with weather type, severity level, time of day, and sensor-lens conditions, enabling attribute-aware evaluation and controlled robustness analysis. The resolution of the dataset is 1242 × 375 for training and 1216 × 352 for testing. The dataset includes eight compound conditions (day/night × clear/fog/rain/snow), with three severity levels for each degraded weather in three driving scenarios (i.e., city, residential, and road). In addition to the weather pairs, we provide matched clean references for a large subset of samples, which supports clean-prior distillation and paired evaluation under controlled degradation. More details on the data distribution across splits and conditions are summarized in Table 1 of Appendix C. Unless otherwise stated, we subsample the validation and test sets to 1,600 frames each for fast ablations and model selection, while keeping the full training set unchanged. We further validate the realism of the simulated RGB and LiDAR degradations by comparing AURORA-KITTI to the real-world adverse-weather dataset MUSES [13] in Fig. 4 and 9. More details of the data realism analysis are discussed in Sec. 5.1.1.

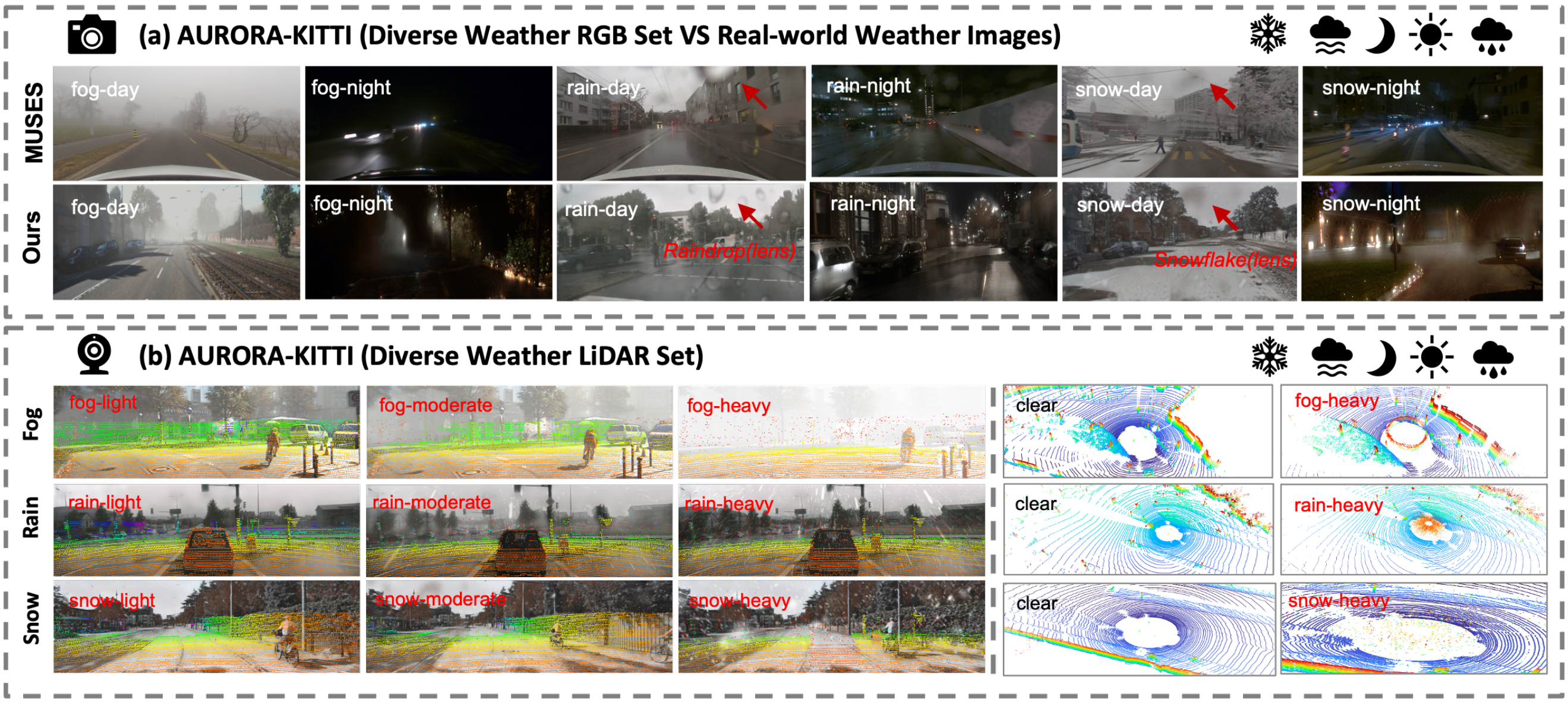

**Figure 4: (a) AURORA-KITTI vs. real adverse weather (MUSES) [13].** Comparison across day/night and fog/rain/snow conditions. AURORA-KITTI reproduces characteristic visibility degradation and lens contamination patterns observed in real driving. **(b) Physics-consistent cross-modal weather severity.** Examples of RGBL pairs under matched light/moderate/heavy weather severity.

***Design Rationale.*** Across rain, fog, and snow, AURORA-KITTI is designed around three principles: (i) *physical controllability*, where severity parameters correspond to measurable quantities (e.g., rain/snow rate, fog attenuation); (ii) *sensor realism*, capturing characteristic artifacts such as attenuation, lost points, range shifts, and weather-induced clutter directly in the raw 3D point cloud; and (iii) *cross-modal consistency*, where RGB and LiDAR are degraded under a perfectly matching weather type and severity for the same scene. Compared to heuristic noise injection, this design enables controlled robustness evaluation and principled study of multi-sensor fusion under adverse conditions. Compared to WeatherKITTI [11], AURORA-KITTI provides: **1)** *Richer degradations on both sensor data*: three severity levels with explicit physical interpretation on both RGB and LiDAR; **2)** *Richer illumination variety*: nighttime and compound conditions (e.g., night-rain, night-snow) which affect the sensing of driving in the dark [40]; and **3)** *Lens occlusions*: in addition to in-air particles, we design adherent simulation models for raindrops and snowflakes that occlude the camera lens, a common phenomenon in outdoor sensing conditions.

### 3.4. Attribute Annotation and Benchmark Protocol

To systematically evaluate the current depth reconstruction models under the proposed DCD setting, we establish a unified benchmarking protocol built upon AURORA-KITTI.

*Unified DCD setting.* All models are evaluated under the same task formulation defined in Eq. (3). Given corrupted multi-modal inputs $(I^w, S^w)$, the objective is to predict the corresponding clean dense ground-truth depth $D^c$. For monocular depth models that do not accept sparse depth, we evaluate them using $I^w$ only. For depth completion models, both $I^w$ (RGB) and $S^w$ (LiDAR) are provided. We consider two evaluation regimes: **(1)** direct zero-shot evaluation. Publicly available pre-trained checkpoints (typically trained on clear-weather datasets) are directly tested on AURORA-KITTI without adaptation. This quantifies the domain gap induced by adverse weather. **(2)** in-domain retraining. Sparse2Dense and BP-Net are retrained on the AURORA-KITTI training split under the DCD objective. Importantly, architectural designs remain unchanged; only the training data domain is altered. The before/after comparison for each fixed architecture estimates the effect of weather-aware training. The BP-Net-based DDCD component ablation separately examines the additional effect of distillation under the same in-domain data; the zero-shot benchmark alone does not isolate method-design effects.

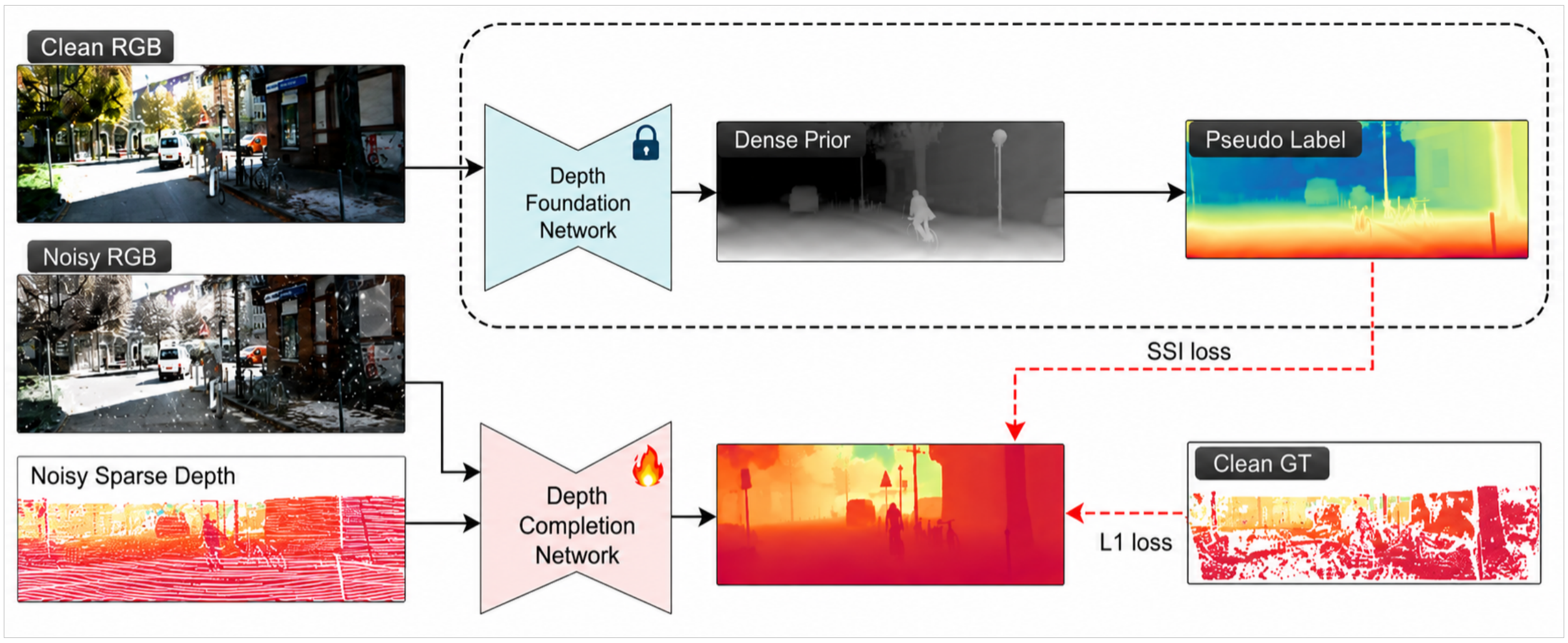

**Figure 5: The proposed clean structural prior distillation pipeline (DDCD).** During training, a frozen depth foundation teacher predicts a dense prior from the paired clean RGB image. After GT-aligned normalisation, this prior provides structural guidance to the DCD network via scale-and-shift invariant (SSI) distillation. The student is supervised jointly by sparse, clean ground-truth data and the aligned teacher prior. The teacher is used only during training. The supervised L1 loss is computed only on valid ground-truth pixels, unlabelled pixels are excluded.

*Cross-dataset generalisation.* To assess whether robustness learned from synthetic adverse weather transfers to real-world conditions, we additionally evaluate retrained models on an external real-world benchmark (DENSE). This measures whether training on AURORA-KITTI improves out-of-domain generalisation without architectural modification.

### 3.5. Proposed Baseline: DDCD

To provide a strong and reproducible reference baseline for AURORA-KITTI, we introduce DDCD, a lightweight distillation-based pipeline built on existing RGB–LiDAR depth-completion backbones. During training, a frozen depth foundation teacher predicts a dense structural prior from the paired clean RGB image. This prior is aligned to the metric ground truth on valid pixels and used as an auxiliary structural supervision together with the standard depth-completion loss. The clean RGB image and teacher model are used only during training; at inference, DDCD receives the same inputs as other DCD models, namely the weather-corrupted RGB image and sparse LiDAR depth. This design demonstrates how the paired clean references in AURORA-KITTI can support clean-prior learning without changing the inference setting.

Under severe weather corruption, fine structures (e.g., thin objects, sharp boundaries) are often lost in $(I^w, S^w)$, even though clear-weather metric depth labels $D^c$ provide metric constraints. In addition, the input raw depth measurements are inherently sparse, failing to provide the fine-grained structural information required for dense, pixel-wise supervision. To enhance structural fidelity, we distil dense geometric priors from a frozen depth foundation teacher into the DCD network.

**1) Teacher prior normalisation and construction.** Given a paired clear-weather RGB image $I^c$, we obtain a dense teacher prediction

$$P = f_\phi(I^c), \tag{5}$$

where the frozen foundation teacher $f_\phi$ may output either a disparity-like relative prediction or a metric depth prediction. Let $D^c$ denote the clear-weather metric depth ground truth and let $M(\mathbf{x}) \in \{0, 1\}$ indicate pixels with valid depth measurements. Since different teachers predict different geometric quantities, a single normalisation is insufficient. We therefore perform image-wise scale-and-shift fitting on valid pixels to convert the teacher output into a supervision-compatible dense prior $D^t$. For disparity-like teachers, we first fit in the inverse-depth domain: $\frac{1}{D^c(\mathbf{x})} \approx a\,P(\mathbf{x}) + b$, $\forall \mathbf{x}$ s.t. $M(\mathbf{x}) = 1$, and obtain the metric teacher prior as

$$D^t(\mathbf{x}) = \frac{1}{a\,P(\mathbf{x}) + b}. \tag{6}$$

For metric-depth teachers, we instead fit $D^c(\mathbf{x}) \approx a\,P(\mathbf{x}) + b$ on valid pixels and set $D^t(\mathbf{x}) = a\,P(\mathbf{x}) + b$. This design preserves the teacher's structural geometry while correcting global scale and shift bias. The teacher remains frozen and is used only during training.

**2) Scale-and-shift invariant (SSI) distillation.** Although $D^t$ is GT-aligned, residual global mismatch across scenes and teachers may remain. Our SSI distillation follows the multi-scale scale-and-shift-invariant matching formulation of [26], while the DCD-specific design lies in the GT-aligned teacher normalisation and the use of paired clean references enabled by AURORA-KITTI.

Let $l \in \{0, 1, \ldots, L-1\}$ index the pyramid levels of the student prediction, and let $\hat{D}^{s,(l)}$ denote the student output at level $l$. We resize the teacher prior to the same spatial resolution: $D^{t,(l)} = \text{Down}_l(D^t)$, where $\text{Down}_l(\cdot)$ denotes resizing and downsampling to match the size of $\hat{D}^{s,(l)}$. We then align $D^{t,(l)}$ to $\hat{D}^{s,(l)}$ via an optimal affine transformation. Notably, $\hat{D}^{s,(l)}$ serves as the reference to estimate $(\alpha_l^*, \beta_l^*)$, ensuring invariance to global scale and shift:

$$(\alpha_l^*, \beta_l^*) = \arg\min_{\alpha,\beta} \left\| W^{(l)} \odot \left( \hat{D}^{s,(l)} - (\alpha D^{t,(l)} + \beta) \right) \right\|_2^2, \quad (7)$$

where $W^{(l)}$ is an optional validity mask (e.g., downsampled from $M$ when available; otherwise set to all-ones), and $\odot$ denotes element-wise masking. We denote the aligned teacher prior as

$$\tilde{D}^{t,(l)} = \alpha_l^* D^{t,(l)} + \beta_l^*. \quad (8)$$

The SSI distillation loss is then defined as

$$\mathcal{L}_{\text{ssi}} = \sum_{l=0}^{L-1} \delta_l \left\| W^{(l)} \odot \left( \hat{D}^{s,(l)} - \tilde{D}^{t,(l)} \right) \right\|_1, \quad (9)$$

where $\{\delta_l\}$ are fixed multi-scale weights.

**3) Residual gradient regularisation.** While SSI distillation enforces global structural consistency, heavy corruption may still induce local high-frequency artefacts. Instead of smoothing depth directly, we regularise the spatial gradients of the distillation residual:

$$R^{(l)} = \hat{D}^{s,(l)} - \tilde{D}^{t,(l)}, \quad (10)$$

where $R^{(l)}$ is the per-pixel difference between the student prediction and the aligned teacher prior, i.e., the same error term appearing inside $\mathcal{L}_{\text{ssi}}$. We apply gradient regularisation on $R^{(l)}$:

$$\mathcal{L}_{\text{grad}} = \sum_{l=0}^{L-1} \delta_l \left( \|\nabla_x R^{(l)}\|_1 + \|\nabla_y R^{(l)}\|_1 \right). \quad (11)$$

In practice, we compute gradients at multiple resolutions by progressively downsampling the residual, which stabilises training under severe weather corruption while preserving depth discontinuities.

*Overall objective.* The supervised term is a masked $L_1$ loss, computed only on pixels where the clean ground truth $D^c$ is valid:

$$\mathcal{L}_{\text{sup}} = \frac{\sum_{\mathbf{x}} M(\mathbf{x}) \left| \hat{D}(\mathbf{x}) - D^c(\mathbf{x}) \right|}{\sum_{\mathbf{x}} M(\mathbf{x}) + \epsilon}, \quad (12)$$

where $M(\mathbf{x}) \in \{0, 1\}$ is the validity mask. Invalid or unlabelled pixels are ignored and are never treated as zero-depth supervision; for multi-scale predictions, the same masking is applied after resizing the target and mask to each prediction scale. The final training objective combines clean-depth supervision, SSI distillation, and residual gradient regularisation:

$$\mathcal{L} = \mathcal{L}_{\text{sup}} + \lambda_d \mathcal{L}_{\text{ssi}} + \lambda_g \mathcal{L}_{\text{grad}}, \quad (13)$$

where $\lambda_d$ and $\lambda_g$ are positive weights. $\mathcal{L}_{\text{sup}}$ enforces metric correctness with $D^c$, $\mathcal{L}_{\text{ssi}}$ transfers clean structural knowledge from $D^t$ in a scale- and shift-invariant manner, and $\mathcal{L}_{\text{grad}}$ suppresses local artefacts. The teacher prior is used for pseudo-guidance only during training.

### 3.6. LiDAR-aware Denoising Metrics for DCD

Standard depth-completion metrics (RMSE, MAE, iRMSE, iMAE) measure the global discrepancy between the predicted and ground-truth depth, but cannot tell whether an error stems from ordinary completion failure or from the propagation of weather-induced LiDAR corruption. This distinction is central to DCD: under adverse weather, the sparse input is not merely incomplete but may contain physically invalid returns from back-scattering, attenuation, and range perturbation. A robust model should therefore (i) reject corrupted returns, (ii) recover regions where valid returns are erased, and (iii) preserve reliable measurements. We introduce three LiDAR-aware denoising metrics that isolate these behaviours: the *Error-Reduction Rate* (ERR $\uparrow$) for corrupted-return correction, the *Dropout Completion Error* (DCE $\downarrow$) for dropout recovery, and the *Clean-Point Preservation* error (CPP $\downarrow$) as a data-fidelity control. The paired clean LiDAR is used *only* to define evaluation regions; every error is computed against the dense ground truth, so the metrics are model-agnostic and can be computed post-hoc for any predicted depth map.

*Notation.* Let $P$, $G$, $S_c$, and $S_w$ denote the predicted dense depth, the dense ground truth, and the clean and weather-corrupted sparse LiDAR projections, with valid masks

$$M_g = \mathbb{1}[G > 0], \quad M_c = \mathbb{1}[S_c > 0], \quad M_w = \mathbb{1}[S_w > 0], \quad (14)$$

where $\mathbb{1}[\cdot]$ is the indicator function. Using the clean–weather pair $(S_c, S_w)$, every sparse pixel falls into exactly one of four noise classes:

$$\begin{aligned} M_{\text{drop}} &= M_c \wedge \neg M_w, \\ M_{\text{spur}} &= M_w \wedge \neg M_c, \\ M_{\text{pert}} &= M_c \wedge M_w \wedge \left( |S_c - S_w| > \tau S_c \right), \\ M_{\text{keep}} &= M_c \wedge M_w \wedge \left( |S_c - S_w| \le \tau S_c \right). \end{aligned} \quad (15)$$

i.e. dropped clean returns, false weather returns, surviving-but-corrupted returns, and reliably preserved returns. We set $\tau = 0.05$, a conservative 5% range tolerance. Intersecting these masks with the valid ground-truth region gives the evaluation regions

$$\begin{aligned} \Omega_{\text{cor}} &= (M_{\text{spur}} \vee M_{\text{pert}}) \wedge M_g, \\ \Omega_{\text{drop}} &= M_{\text{drop}} \wedge M_g, \\ \Omega_{\text{keep}} &= M_{\text{keep}} \wedge M_g. \end{aligned} \quad (16)$$

These cover, respectively, pixels where the weather return exists but is unreliable, where a valid return is erased, and where the return remains reliable. Clear frames yield empty noise regions, a built-in sanity check. For any region $\Omega$, the regional RMSE against the ground truth is calculated.

**Table 3**
Benchmarking results on the AURORA-KITTI test set. DDCD is trained in-domain with a paired clean-RGB teacher during training; the other listed models use off-the-shelf zero-shot checkpoints. Their performance gap therefore reflects different training protocols as well as model designs. Row colours denote the input modality: LiDAR-only , RGB-only , and RGB+LiDAR .

| Method | Year | Setting | RMSE↓ | MAE↓ | iRMSE↓ | iMAE↓ | ERR↑ | DCE↓ | CPP↓ | Params | Time |
|---|---|---|---|---|---|---|---|---|---|---|---|
| IPBasic(fast)[41] | CRV'18 | Zero-shot | 7861.15 | 2650.48 | 92.55 | 31.64 | -0.001 | 2416.93 | 1810.5 | None | 16.4 |
| IPBasic(multi)[41] | CRV'18 | Zero-shot | 6916.28 | 2209.20 | 43.49 | 14.34 | 0.094 | 2103.64 | 1998.3 | None | 29.70 |
| S2Dense(SS)[5] | ICRA'19 | Zero-shot | 5971.83 | 2221.87 | 23.17 | 10.74 | 0.051 | 1850.15 | 1860.9 | 26.1M | 6.1 |
| S2Dense(Su)[5] | ICRA'19 | Zero-shot | 4660.44 | 1610.71 | 18.52 | 7.38 | 0.252 | 1609.60 | 1082.0 | 26.1M | 6.1 |
| NLSPN [17] | ECCV'20 | Zero-shot | 4013.23 | 1293.62 | 19.86 | 5.15 | 0.071 | 1375.59 | 1053.8 | 26.2M | 40.2 |
| CFormer [18] | CVPR'23 | Zero-shot | 3961.99 | 1265.66 | 14.15 | 3.77 | 0.164 | 1315.77 | 1029.6 | 83.5 M | 73.4 |
| BP-Net [7] | CVPR'24 | Zero-shot | 4945.43 | 1818.58 | 12.32 | 5.73 | 0.348 | 1466.38 | 1663.8 | 89.9M | 48.6 |
| Marigold-DC[33] | ICCV'25 | Zero-shot | 6778.75 | 2441.68 | 54.67 | 16.58 | 0.147 | 2106.19 | 1843.7 | - | 34.0*s* |
| OMNI-DC [20] | ICCV'25 | Zero-shot | 6225.81 | 2080.96 | 42.34 | 15.05 | 0.010 | 2089.78 | 1415.7 | 420.7M | 320 *ms* |
| DMD3C [26] | CVPR'25 | Zero-shot | 7244.44 | 2638.07 | 42.65 | 15.86 | 0.003 | 2241.20 | 1826.3 | 89.87M | 48.60 |
| UniK3D[42] | CVPR'25 | Zero-shot | 3716.06 | 1552.24 | 9.37 | 6.36 | 0.744 | 683.03 | 3136.4 | 358.8 M | 45.1 |
| DA-V3[25] | ICLR'26 | Zero-shot | 4564.44 | 2903.61 | 23.81 | 20.53 | 0.683 | 888.20 | 4315.2 | 1689.8M | 165.7 |
| DDCD **(Ours)** | 2026 | **In-domain** | **1801.93** | **570.36** | **4.25** | **1.96** | 0.826 | 474.41 | 835.1 | 89.87M | 48.60 |

*Error-Reduction Rate (ERR↑).* Spurious and range-perturbed returns are outliers in the sparse input; a robust model should lower their residual error relative to the corrupted input itself:

$$\mathrm{ERR} = 1 - \frac{\mathrm{RMSE}_{\Omega_{\mathrm{cor}}}(P)}{\mathrm{RMSE}_{\Omega_{\mathrm{cor}}}(S_w)}. \tag{17}$$

ERR = 1 denotes perfect correction, ERR = 0 a prediction no better than the corrupted input, and ERR < 0 amplification of the corruption.

*Dropout Completion Error (DCE↓).* Adverse weather also erases valid returns. With the anchor missing, the model cannot reject an observed outlier and must instead infer geometry from neighbouring LiDAR, RGB appearance, and learned priors. We evaluate this directly on the dropout region:

$$\mathrm{DCE} = \mathrm{RMSE}_{\Omega_{\mathrm{drop}}}(P). \tag{18}$$

ERR and DCE thus capture the two principal failure modes of adverse-weather LiDAR: erroneous returns and missing returns.

*Clean-Point Preservation (CPP↓).* A high denoising score must not be obtained by discarding the sparse input wholesale. Reliable returns are strong geometric anchors, so we measure prediction error where the return is present and consistent with the clean reference:

$$\mathrm{CPP} = \mathrm{RMSE}_{\Omega_{\mathrm{keep}}}(P). \tag{19}$$

Overall, an effective DCD model should therefore attain high ERR, low DCE, and low CPP simultaneously. Since paired clean and weather-corrupted sparse LiDAR is needed to calculate the LiDAR-aware denoising metrics, these metrics are only reported on AURORA-KITTI.

**Table 4**
Benchmarking results on the real-world DENSE test set. Lower values are better. DDCD was trained on AURORA-KITTI with paired clean-RGB teacher supervision; the other listed methods are evaluated zero-shot. These rows are not a matched-training comparison. Row colours denote the input modality: LiDAR-only , RGB-only , and RGB+LiDAR .

| Method | Setting | RMSE↓ | MAE↓ | iRMSE↓ | iMAE↓ |
|---|---|---|---|---|---|
| IPBasic(F)[41] | Zero-shot | 7830.63 | 5582.13 | 286.58 | 217.22 |
| IPBasic(M)[41] | Zero-shot | 7346.78 | 5172.20 | 176.23 | 116.17 |
| S2Dense(SS)[5] | Zero-shot | 5859.08 | 3916.45 | 54.91 | 34.76 |
| S2Dense(Su)[5] | Zero-shot | 5416.31 | 3671.62 | 99.12 | 64.14 |
| NLSPN[17] | Zero-shot | 4916.51 | 3223.92 | 71.12 | 45.13 |
| CFormer[18] | Zero-shot | 4754.04 | 3234.06 | 50.43 | 28.58 |
| BP-Net[7] | Zero-shot | 4946.62 | 3561.48 | 177.36 | 85.51 |
| MarigoldDC[33] | Zero-shot | 5364.58 | 3882.82 | 130.38 | 92.30 |
| OMNI-DC[20] | Zero-shot | 5752.49 | 4031.82 | 189.41 | 140.97 |
| DMD3C[26] | Zero-shot | 6042.42 | 4223.60 | 184.72 | 130.57 |
| DA-V3[25] | Zero-shot | 7778.43 | 6118.93 | 120.46 | 83.00 |
| UniK3D[42] | Zero-shot | 3509.04 | 2485.10 | 49.06 | 30.77 |
| DDCD **(Ours)** | **In-domain** | **2761.97** | **1569.14** | **34.82** | **13.24** |

# 4. Experimental Setup

*Benchmark Models.* We evaluate ten representative depth estimation and depth completion approaches, including classical models, monocular depth foundation models, and unsupervised and supervised paradigms. Specifically, these include: (i) IP-Basic [41] as a classical LiDAR-only completion baseline; (ii) Depth Anything V3 [25] as a large-scale monocular depth foundation models; (iii) Sparse2Dense [5] (unsupervised and supervised variants); (iv) BP-Net [7] and DMD3C [26] as supervised RGB–LiDAR depth completion models; (v) NLSPN [17] and Completion-Former [18] as spatial-propagation- and transformer-based RGB–LiDAR completion models; (vi) OMNI-DC [20] as a robustness-oriented multi-resolution completion method; (vii) Marigold-DC [33] as a diffusion-based zero-shot method; and (viii) UniK3D [42] as a universal metric depth estimation model for outdoor driving scenes. Together, these baselines cover LiDAR-only, monocular depth estimation,

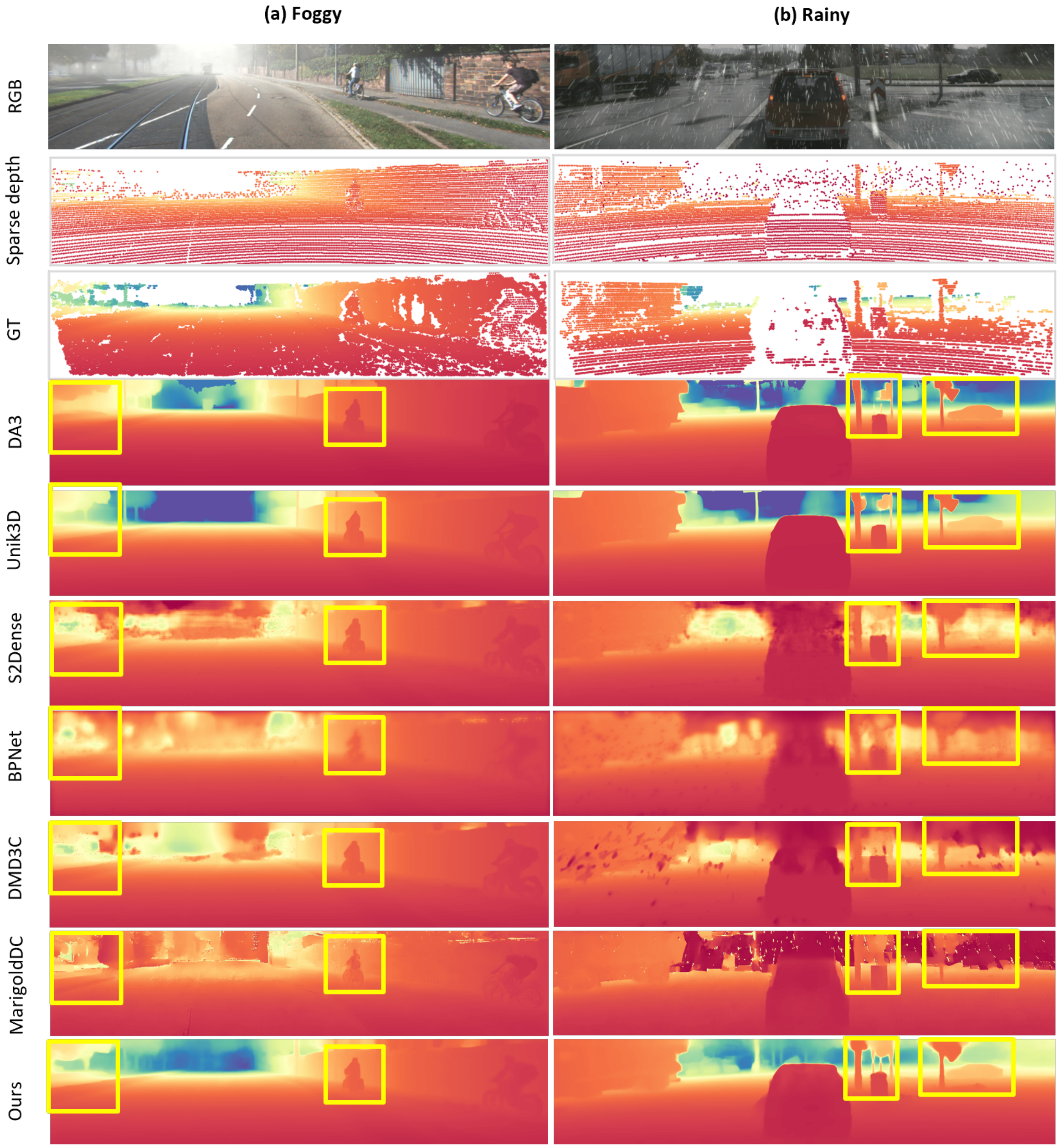

**Figure 6:** The benchmarking qualitative visualisation results on the AURORA-KITTI dataset under foggy and rainy conditions.

and RGB–LiDAR fusion paradigms from classical to recent methods.

*Datasets.* We conduct a comprehensive evaluation on two benchmarks: **AURORA-KITTI** is our proposed large-scale synthetic adverse-weather RGB–LiDAR benchmark with paired clean depth supervision, diverse illumination, weather conditions, and matched severity controls across modalities (Sec. 3.2). **DENSE: Pixel-Accurate Depth Benchmark** is used to assess cross-dataset generalisation under real-world adverse weather. It provides high-resolution, per-pixel metric depth references with angular resolution comparable to a 50 Mp camera, enabling fine-grained evaluation for depth estimation and completion beyond sparse LiDAR ground-truth data used in traditional benchmarks. The benchmark covers four canonical automotive scenes (pedestrian zone, residential area, construction zone, and highway) recorded under clear, rainy, and foggy conditions in both daytime and nighttime, with a wide range of visibility levels for controlled analysis of weather severity [9]. This real-world dataset allows us to evaluate whether robustness learned from synthetic AURORA-KITTI transfers to complex real adverse-weather environments with pixel-accurate depth supervision.

*Evaluation Metrics.* We follow the standard depth completion protocol and report root mean squared error (RMSE), mean absolute error (MAE), and their inverse-depth counterparts, root mean squared error of the inverse depth (iRMSE), and mean absolute error of the inverse depth (iMAE). Beyond standard depth-completion metrics, we introduce three LiDAR-aware denoising metrics, ERR, DCE, and CPP, to jointly assess completion accuracy and weather-noise suppression (see Sec. 3.6).

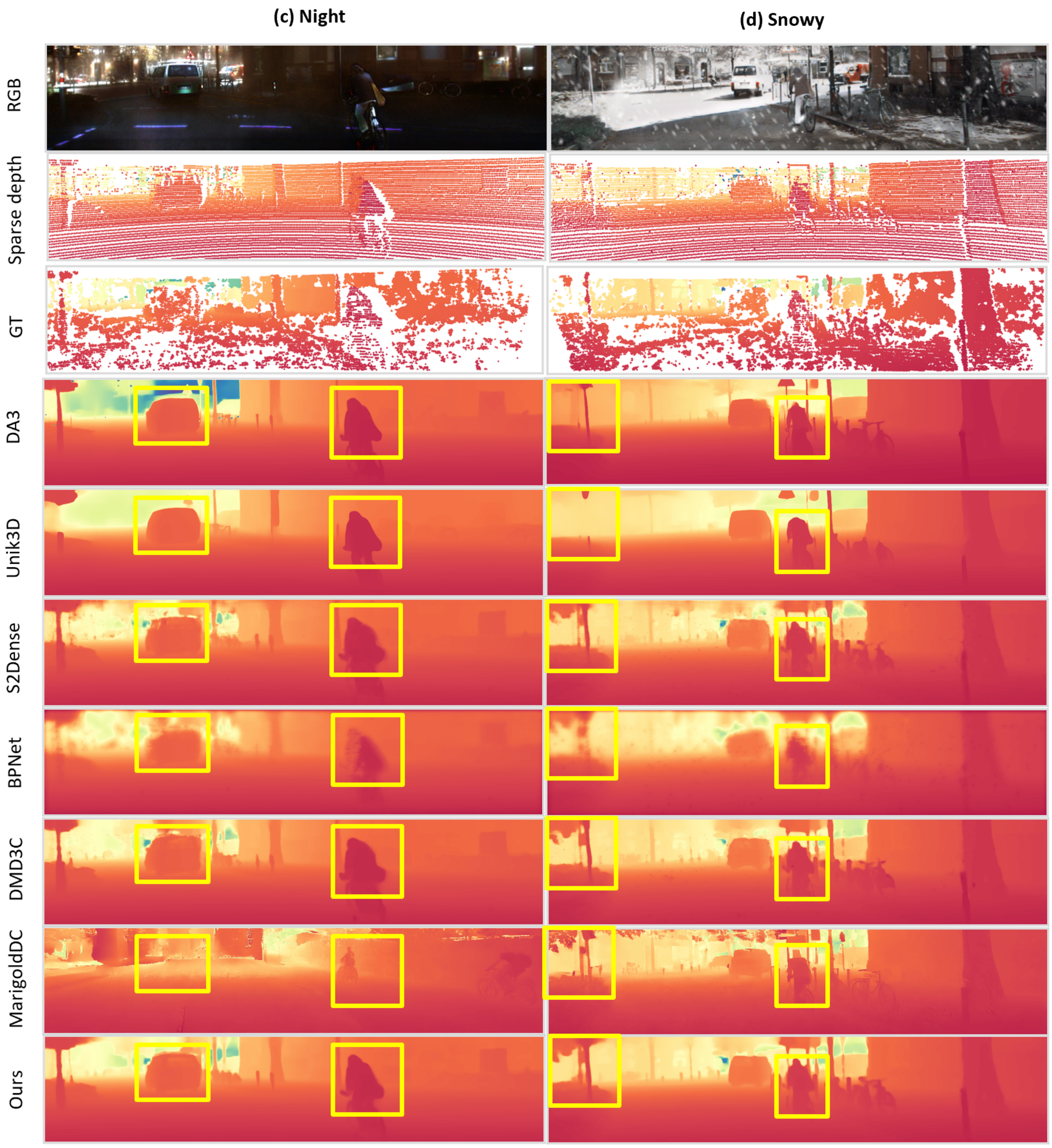

**Figure 7:** The benchmarking qualitative visualisation results on the AURORA-KITTI dataset under night and snowy conditions.

*Implementation Details.* All retraining experiments are conducted on NVIDIA RTX 4090 GPUs. For DDCD, the student is BP-Net [7] by default, initialised from a mixed-domain depth-completion checkpoint and fine-tuned on AURORA-KITTI for 30 epochs. We use AdamW (weight decay 0.05) with a peak learning rate of $5 \times 10^{-4}$, cyclic momentum in [0.85, 0.90], and batch size 2. To enable a single, consistent evaluation protocol across both AURORA-KITTI and the real-world DENSE benchmark, we adopt a unified bottom-crop of $256 \times 1216$ for both training and evaluation, following the standard BP-Net evaluation convention. This lower-region crop covers the depth-valid field of view shared by both datasets and removes resolution mismatch as a confounding factor in cross-dataset comparison. The DDCD objective combines masked clean-depth supervision with scale-and-shift-invariant (SSI) distillation and residual-gradient regularisation, with loss weights $\lambda_d = 0.1$ and $\lambda_g = 0.05$. The denoising-metric tolerance $\tau = 0.05$ is used only for evaluation and does not affect training. Unless otherwise stated, we adopt each method's original optimizer and learning schedule to ensure faithful reproduction, modifying only the training-data domain, thereby isolating the data-domain effect while preserving each method's original architecture and optimisation. The distillation teacher is a frozen Depth

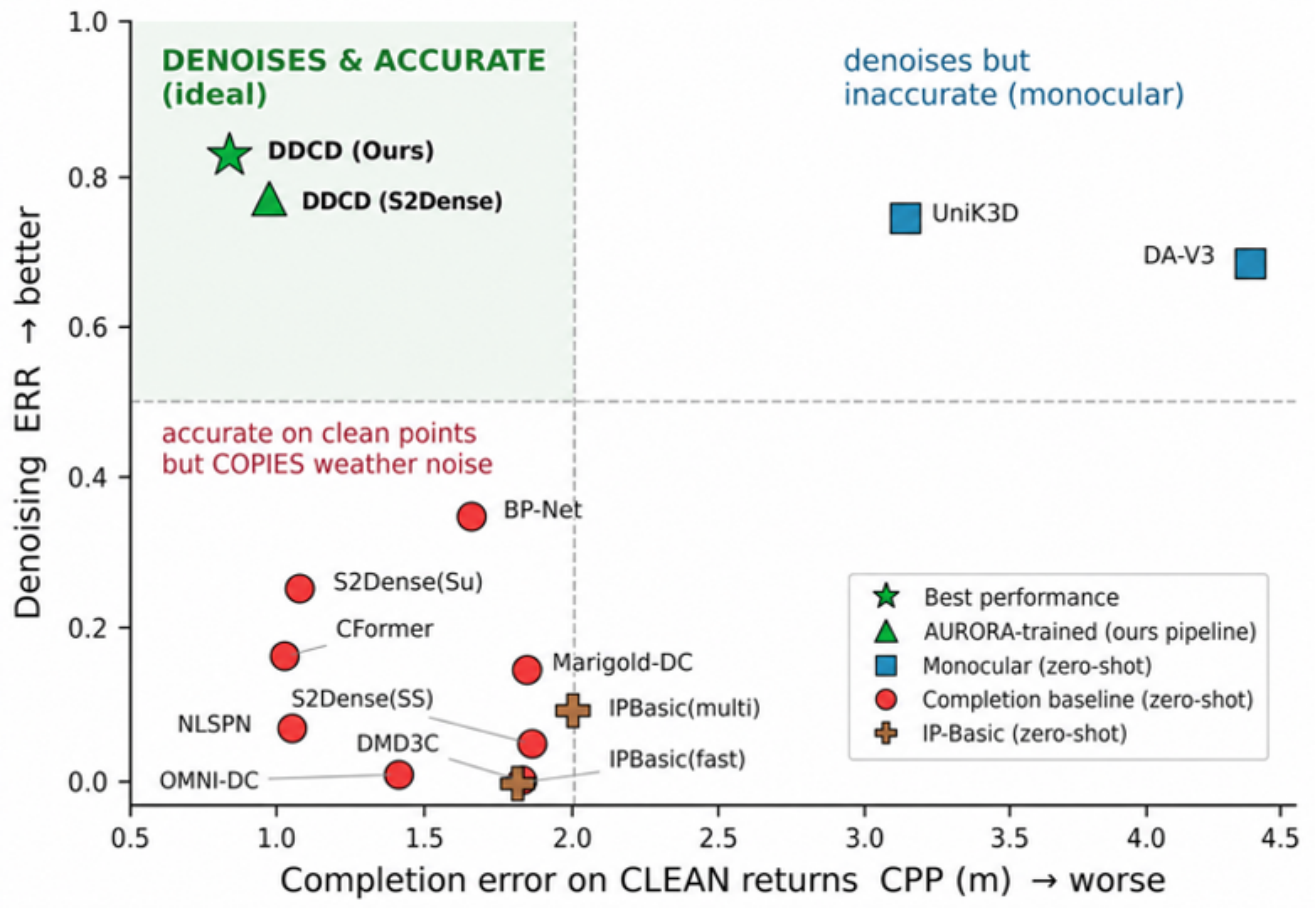

**Figure 8:** Benchmarking visualisation of the proposed denoising metrics in the ERR–CPP plane.

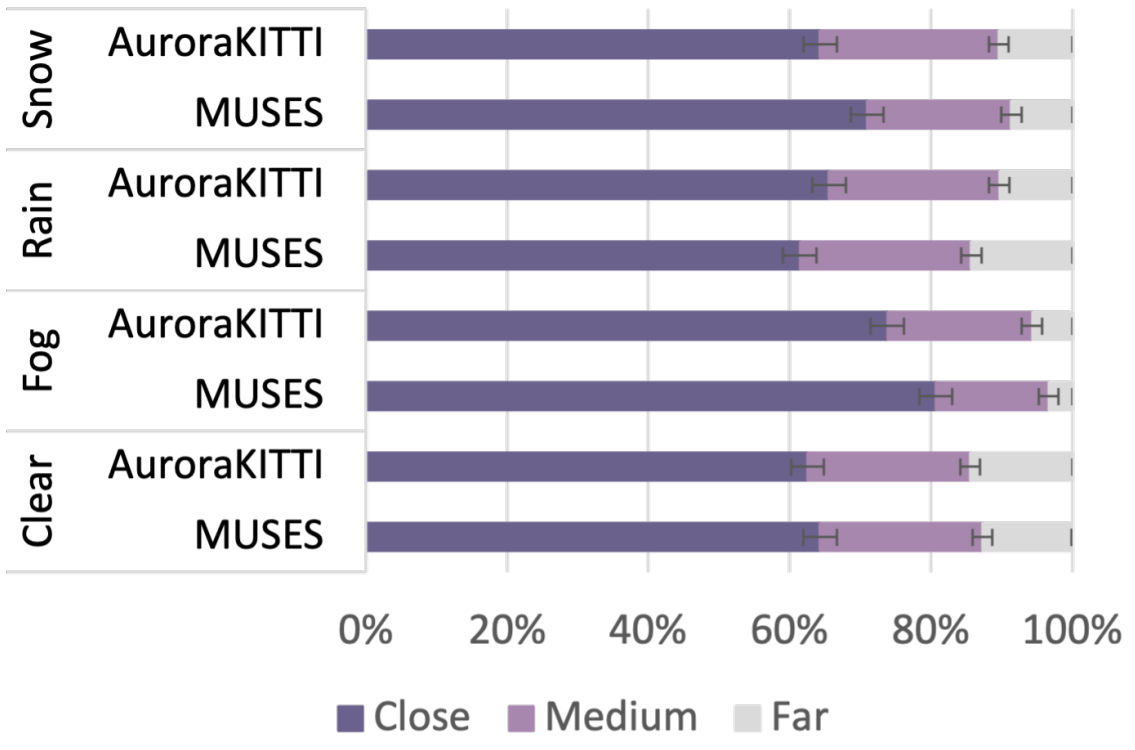

**Figure 9: LiDAR realism under adverse weather.** Range-related statistics of AURORA-KITTI compared with MUSES [13] across weather types, showing consistent effective-range degradation trends.

Anything V2 [24], used only during training. We select it as our reference teacher to supply a dense relative-depth structural prior from the paired clean scene, and prioritise transfer to real adverse weather when choosing between candidate teachers. Valid clean depth pixels align the teacher prior to metric scale before it supervises the student. The V2-versus-V3 experiment later tests the sensitivity of this choice: V3 performs better on AURORA-KITTI, whereas V2 transfers better to DENSE (Table 9). This choice is specific to the evaluated protocol rather than a claim that V2 is universally superior. It is applied to the paired clean RGB image to provide a weather-invariant structural prior, while the student receives only the weather-corrupted RGB image and sparse LiDAR. This is privileged training-only supervision, which can improve performance relative to models that do not have paired clean references. Teacher predictions are precomputed offline, so no teacher forward pass is needed during training, and neither the teacher nor the paired clean RGB is used at inference.

# 5. Experimental Results and Analysis

This section presents the benchmark results and analysis.

## 5.1. Benchmark Results

### *5.1.1. AURORA-KITTI Realism Evaluation*

We evaluate whether AURORA-KITTI captures realistic adverse-weather effects by comparing it with MUSES [13], a real-world multi-modal driving dataset collected under diverse weather.

*Qualitative Results* Fig. 4 (a) shows qualitative RGB comparisons across fog, rain, and snow under both day-time and nighttime conditions. Our renderings reproduce key visual signatures observed in real captures, including contrast attenuation, haze veiling in fog, rain-induced streaks and lens contamination from raindrops and snowflakes. Importantly, these artefacts are spatially non-uniform and scene-dependent, matching the heterogeneous corruption patterns commonly seen in the real world. We further validate LiDAR realism using range-related statistics (Fig. 9). Across weather conditions, AURORA-KITTI exhibits similar effective-range degradation trends to MUSES: dense fog produces the strongest attenuation and volumetric backscatter, leading to a shorter effective sensing range and a higher proportion of near-range returns, while clear weather and rain preserve longer-range measurements comparatively.

*Quantitative realism of the synthetic weather.* Beyond qualitative examples and aggregate range statistics, we quantify the LiDAR degradation directly at the dataset level by applying the four denoising masks of §3.6 ($M_{\text{drop}}, M_{\text{spur}}, M_{\text{pert}}, M_{\text{keep}}$) to the input paired clean/weather sparse projections. Table 5 reports, per weather and severity, the fraction of clean returns lost (dropout), weather-induced false returns (spurious), range-perturbed returns, preserved returns, effective sensing range, and near-range clutter. The characterisation matches the intended physics: clear frames leave the masks empty by construction, fog is the strongest degrader at every severity (heavy fog drops 41.9% of returns and shortens the effective range to 14.5 m), and the degradation increases monotonically with physical severity for fog and rain across all reported statistics. This quantitatively confirms that AURORA-KITTI's severity control is physically consistent rather than a cosmetic label. Snow's severity is encoded more subtly: the globally pooled rates are relatively flat, while far-range dropout and spurious-return counts increase once near-range pixel volume is factored out.

These visual and geometric consistencies in both modalities indicate that AURORA-KITTI provides a realistic, reliable and cross-modally aligned proxy for systematic DCD benchmarking and robustness training.

### *5.1.2. AURORA-KITTI Benchmark Results*

***Quantitative Results.*** Table 3 reports the performance of ten existing representative depth estimation and depth completion methods on the AURORA-KITTI test split (1600 pairs) together with our proposed DDCD method. Lower values indicate better performance for RMSE, MAE, iRMSE, iMAE, DCE, and CPP, while higher ERR indicates

**Table 5**
Dataset-level LiDAR degradation statistics on AURORA-KITTI. Rates are computed from paired clean/weather sparse LiDAR masks. *Statistics are computed over the combined 3,200-frame validation and test subsets.*

| Weather | Severity | $N$ | Dropout(%) | Spurious(%) | Perturbed(%) | Preserved(%) | Effective Range(m) | Near-Clutter(%) |
|---|---|---|---|---|---|---|---|---|
| Clear | – | 763 | 0.00 | 0.00 | 0.00 | 100.00 | 42.85 | 23.42 |
| Fog | Light | 259 | 6.46 | 2.08 | 0.04 | 93.50 | 31.22 | 24.65 |
| | Moderate | 265 | 19.90 | 4.25 | 0.33 | 79.83 | 22.03 | 32.64 |
| | Heavy | 273 | 41.93 | 7.41 | 0.50 | 57.78 | 14.48 | 42.37 |
| Rain | Light | 292 | 8.45 | 0.00 | 0.00 | 91.55 | 41.07 | 24.83 |
| | Moderate | 295 | 20.18 | 2.60 | 0.20 | 79.66 | 26.51 | 28.30 |
| | Heavy | 290 | 21.62 | 3.96 | 0.30 | 78.15 | 25.42 | 30.38 |
| Snow | Light | 251 | 19.78 | 2.13 | 0.26 | 80.01 | 33.06 | 25.51 |
| | Moderate | 254 | 20.61 | 2.63 | 0.36 | 79.11 | 32.64 | 25.32 |
| | Heavy | 258 | 20.39 | 2.89 | 0.40 | 79.29 | 32.09 | 25.44 |

*Note: Effective range is the mean per-frame 95% valid LiDAR range. Near-clutter denotes the fraction of weather-corrupted LiDAR returns within 8 m.*

better corrupted-return correction. Compared to learning-based approaches, classical interpolation baselines (IP basic) remain worse across all metrics, confirming the limited robustness of purely heuristic interpolation strategies under severe degradations.

In contrast, all deep-learning-based approaches perform better than the classical interpolation baselines in adverse conditions. Among lightweight models, the supervised version of Sparse2Dense achieves competitive performance with an RMSE of 4660.44 mm while maintaining very high efficiency (6.1 ms per frame with 26.1M parameters), highlighting a favourable accuracy–efficiency trade-off. Notably, supervised Sparse2Dense now surpasses the heavier fusion models BP-Net (4945.43 mm) and DMD3C (7244.44 mm) despite its far simpler U-Net design. Among all existing completion models, the transformer- and propagation-based CompletionFormer (3961.99 mm) and NLSPN (4013.23 mm) are the strongest, whereas the robustness-oriented OMNI-DC degrades sharply under coupled RGB–LiDAR corruption (6225.81 mm, ERR 0.010). Interestingly, recent monocular depth foundation models, including Depth Anything V3 and UniK3D, achieve strong performance despite using RGB only as the input. This result demonstrates the effectiveness of large-scale pretraining in learning generalisable geometric knowledge. In particular, amongst the existing methods, UniK3D attains the lowest RMSE (3716.06 mm) and CompletionFormer the lowest MAE (1265.66 mm). UniK3D uses substantially fewer parameters (358.8M) and lower inference time than Depth Anything V3, demonstrating better generalisability for outdoor metric depth estimation.

***Quantitative Denoising Results.*** The denoising metrics reveal a failure mode invisible to standard error. Fig. 8 visualises all methods in the ERR–CPP plane, which cleanly separates three behavioural regimes that aggregate RMSE conflates. (i) The lower-left "copies-noise" regime: completion baselines (BP-Net, NLSPN, CompletionFormer, DMD3C, OMNI-DC) attain low CPP on reliable returns but near-zero or negative ERR, i.e. they faithfully preserve clean points yet propagate weather-corrupted returns into the dense output. (ii) The lower-right "denoises-but-inaccurate" regime: RGB-only foundation models (UniK3D, DA-V3) reach high ERR by effectively discarding the LiDAR stream, but incur the largest CPP (3136.4 / 4315.2 mm), confirming that they ignore even trustworthy sparse evidence. (iii) The upper-left "ideal" regime, high ERR and low CPP simultaneously, is reached most clearly by DDCD (ERR 0.826, DCE 474.41 mm, CPP 835.1 mm). This confirms that noise propagation and LiDAR neglect are genuinely distinct axes, and that jointly optimising completion and denoising is necessary to occupy the ideal regime.

As an in-domain reference baseline, DDCD achieves the strongest accuracy and denoising results under the proposed protocol. The other methods in this table use zero-shot checkpoints, so this ranking is a practical cross-protocol comparison rather than evidence that DDCD is superior under matched training. Other AURORA-KITTI-trained baselines and a BP-Net/DDCD within-backbone comparison are summarised in Table 8, with the DDCD component ablation in Table 10. In particular, DDCD reduces RMSE to 1801.93 mm, relative to the strongest zero-shot baseline, UniK3D, this corresponds to 1914.13 mm (51.5% relative reduction), and over the diffusion-based Marigold-DC by 4976.82 mm (73.4%). Similar improvements are observed in MAE, iRMSE, and iMAE. Importantly, DDCD maintains a moderate model size (89.87M parameters) and real-time inference speed (48.6 ms), achieving a favourable balance between accuracy and efficiency.

***Qualitative Results.*** Fig. 6 and Fig. 7 present qualitative comparisons under four adverse weather conditions: fog, rain, night, and snow. Severe weather degrades both modalities: RGB images exhibit reduced visibility, illumination degradation, and weather artefacts such as rain streaks and snow occlusions, while projected sparse LiDAR maps suffer from missing returns at long distances and weather-induced outliers. Foundation RGB-only models, such as Depth Anything V3 and UniK3D, recover rich structural details directly from image cues. However, their metric scale estimation remains unstable, particularly in distant regions where noticeable depth compression and abrupt transitions are observed compared to the ground-truth data. Under snowy conditions, fine structures such as tree branches and cyclist silhouettes are occasionally missed due to occlusion by snowflakes in UniK3D in Fig. 7. Sparse2Dense and BP-Net, although leveraging sparse LiDAR input, exhibit noisy boundaries and blurred object contours, suggesting that corrupted LiDAR measurements propagate noise through the fusion process. DMD3C and Marigold-DC preserve object relative structures more faithfully, benefiting from large-scale pretrained representations. Nevertheless, both methods struggle in sky regions and distant areas. It is also worth noting that in rainy scenes, visible rain-streak-like artefacts appear in the predictions of DMD3C and Marigold-DC in Fig. 6, indicating vulnerability to structured weather noise. This qualitative artefact is consistent with Marigold-DC's high quantitative error under rain and fog (see Table 3), jointly showing that diffusion priors trained on clean imagery do not transfer to weather-corrupted inputs.

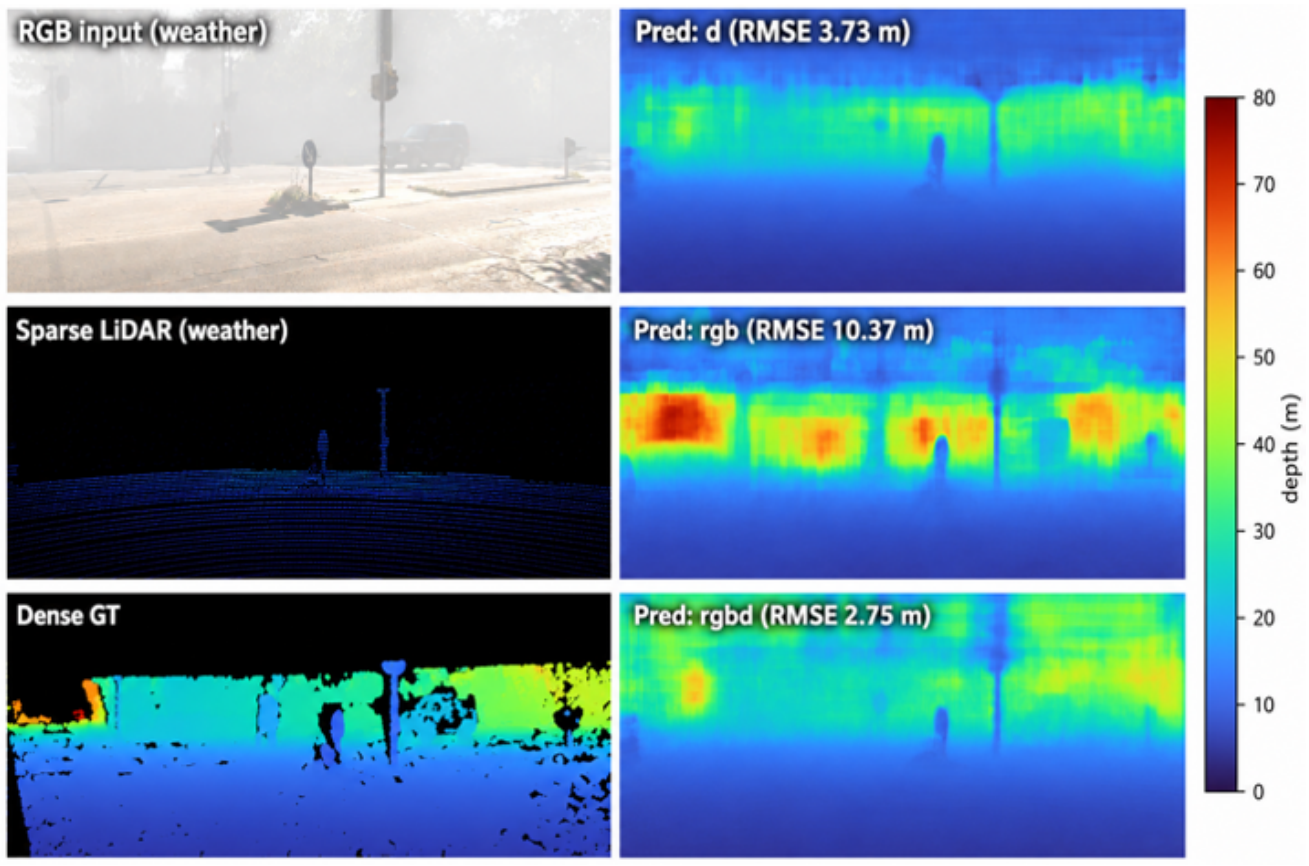

**Figure 10:** Visualisation of results with the modality ablation.

Overall, DDCD produces more consistent scale alignment, sharper object boundaries, and improved robustness to adverse weather artefacts. Distant regions remain geometrically coherent, and fine structures such as vehicles, pedestrians, and trees are better preserved across all evaluated conditions.

***Modality Ablation Study.*** To examine whether both degraded input modalities are necessary for DCD, we conduct a controlled modality ablation with three input settings: RGB image only (**RGB**), sparse depth only (**Depth**), and full RGB-D input (**RGB-D**). Since the propagation-based BP-Net backbone used in our main DDCD model relies on sparse-depth seeds, it is not directly suitable for a LiDAR-free setting. Therefore, we use the concatenation-based Sparse-to-Dense backbone [5], where either RGB or sparse depth can be removed in a well-defined manner. All variants follow the same training and distillation pipeline, and only the student input modality is changed.

Table 6 shows that the full RGB-D model achieves the best performance across both standard depth-completion metrics and LiDAR-aware denoising metrics. Compared with RGB only, RGB-D reduces RMSE from 4623.7 mm to 2188.6 mm and CPP from 3880.4 mm to 1120.8 mm, confirming that sparse LiDAR provides essential metric-scale anchors. Without sparse depth, the RGB-only model relies solely on appearance cues and suffers from large absolute-depth errors. The depth-only variant is substantially stronger than RGB-only, reducing RMSE to 2468.9 mm and CPP to 1223.9 mm. This is because it combines sparse metric-depth anchors with RGB-derived structural supervision from the monocular teacher, even though RGB is not used as the student input. Nevertheless, depth only remains inferior to full RGB-D. Adding RGB to sparse depth further improves ERR from 0.759 to 0.799, and reduces CPP from 1223.9 mm to 1120.8 mm, showing that RGB provides complementary contextual cues for suppressing corrupted LiDAR returns and recovering weather-induced dropout regions. The qualitative comparison in Fig. 10 is consistent with these quantitative results.

**Table 6**
Modality ablation on the AURORA-KITTI dataset using the DDCD (S2Dense) with three modalities: RGB-only, depth-only, and RGB-D inputs.

| Input | RMSE↓ | MAE↓ | iRMSE↓ | iMAE↓ | ERR↑ | CPP↓ |
|---|---|---|---|---|---|---|
| **RGB** only | 4623.7 | 2275.1 | 13.63 | 8.56 | 0.704 | 3880.4 |
| **Depth** only | 2468.9 | 797.2 | 5.77 | 2.56 | 0.759 | 1223.9 |
| **RGB-D** both | **2188.6** | **730.0** | **5.26** | **2.38** | **0.799** | **1120.8** |

Overall, this ablation demonstrates that RGB and sparse LiDAR play complementary roles in DCD: sparse LiDAR anchors metric geometry, while RGB provides contextual priors that improve corrupted-return correction and dropout recovery. This supports our formulation of DCD as a multi-modal task requiring both completion and denoising.

### *5.1.3. DENSE Benchmark Results*

Table 4 reports benchmarking results on the real-world DENSE Pixel-Accurate dataset. Compared with the AURORA-KITTI test set, all methods exhibit consistently higher errors. DENSE contains severe real weather, including fog with visibility ranging from partial degradation to near-zero visibility, and therefore introduces strong scattering effects. It also does not contain snow, creating a domain gap relative to the full synthetic weather space in AURORA-KITTI. A similar overall ranking is observed on DENSE. Classical interpolation methods remain highly sensitive to noise and sparsity corruption. Several RGB–LiDAR completion methods also degrade substantially, especially when their fusion process assumes reliable sparse-depth input. UniK3D shows comparatively stable performance among existing methods, suggesting stronger cross-domain generalisation capability. Nevertheless, DDCD, which is trained on AURORA-KITTI and is therefore not a zero-shot method, obtains an RMSE of 2761.97 mm and serves as the strongest in-domain reference under this protocol. DDCD also achieves the lowest iRMSE and iMAE, indicating superior inverse-depth accuracy, which is particularly important for near-range perception under degraded visibility. These results demonstrate transfer of the AURORA-KITTI-trained DDCD model to real adverse weather; the zero-shot ranking does not by itself isolate the benefit of DDCD from the benefit of its training data. The AURORA-KITTI-retrained Sparse2Dense and BP-Net results are shown alongside DDCD in Table 8.

**Table 7**
Effect of AURORA-KITTI retraining on AURORA-KITTI (top) and DENSE (bottom). "w/r" denotes with retraining.

| Method | w/r | RMSE↓ | MAE↓ | iRMSE↓ | iMAE↓ |
|---|---|---|---|---|---|
| S2Dense | | 4660.44 | 1610.71 | 18.52 | 7.38 |
| S2Dense | ✓ | 2497.56 | 833.34 | 5.27 | 2.45 |
| BP-Net | | 4945.43 | 1818.58 | 12.32 | 5.73 |
| BP-Net | ✓ | 2103.33 | 598.72 | 4.51 | 2.46 |
| S2Dense | | 5416.31 | 3671.62 | 99.12 | 64.14 |
| S2Dense | ✓ | 4973.36 | 3615.53 | 52.74 | 36.95 |
| BP-Net | | 4946.62 | 3561.48 | 177.36 | 85.51 |
| BP-Net | ✓ | 4252.91 | 2375.51 | 42.58 | 22.01 |

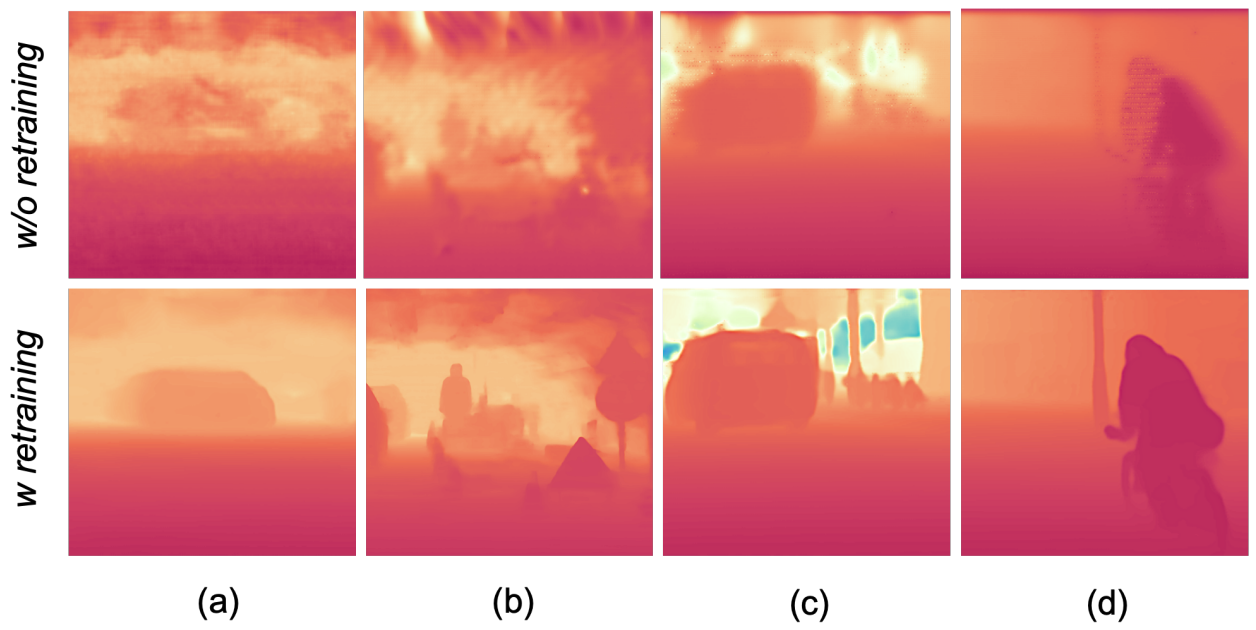

**Figure 11:** Visualisation of results with the pretrained model (top row) or AURORA-KITTI retrained model (bottom row) on BPNet. (a) (b) are examples from the DENSE dataset, and (c) and (d) are from the AURORA-KITTI test set.

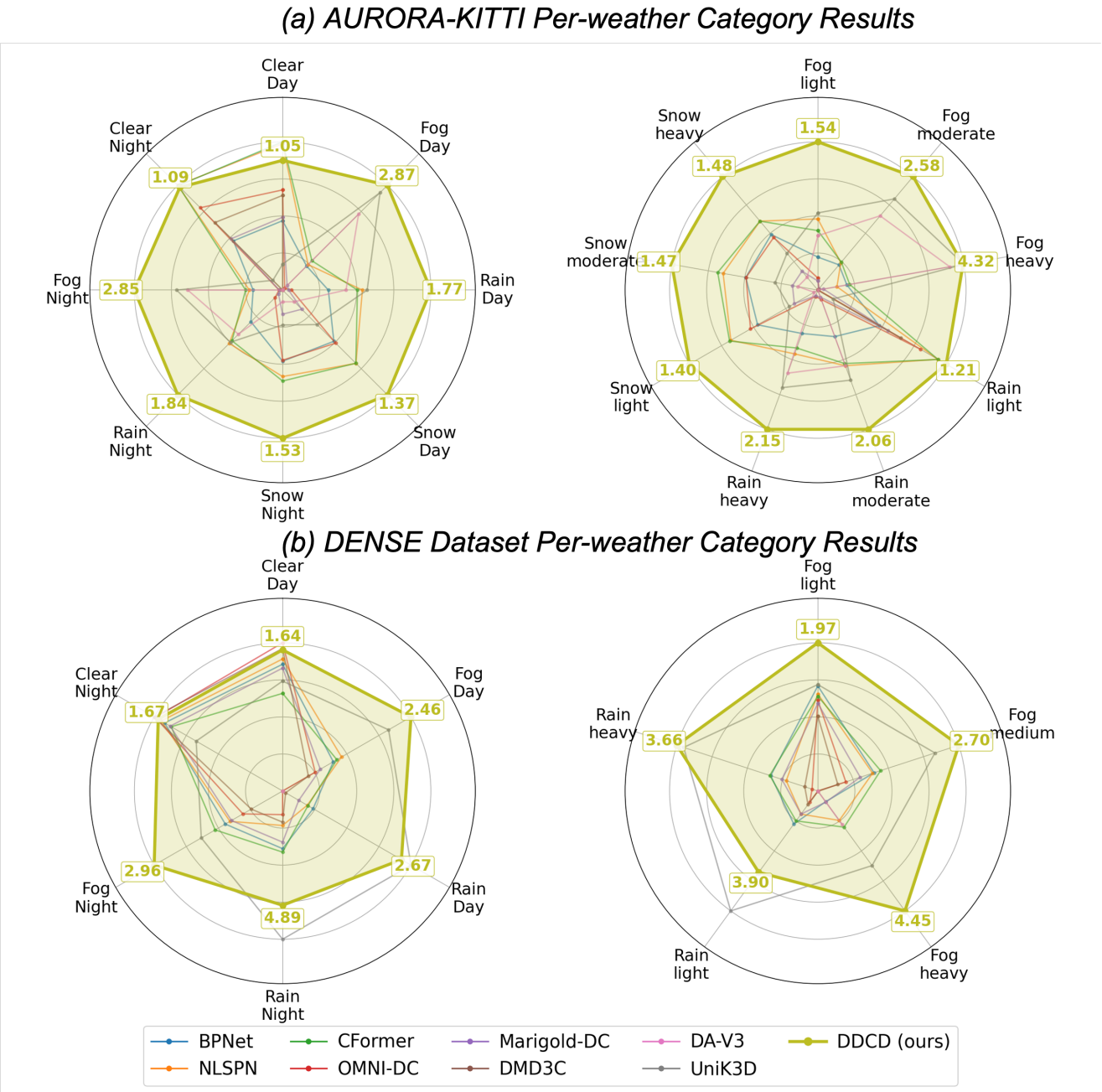

**Figure 12: Per-category RMSE robustness** on (a) synthetic AURORA-KITTI and (b) real-world DENSE. The left chart in each block shows weather×time-of-day, and the right chart shows weather×severity. Each axis is independently normalised and oriented so that lower error lies further out.

### 5.1.4. Robustness under different weather conditions

To provide a finer-grained robustness analysis, we decompose RMSE by *weather type*, *severity*, and *time of day* on both AURORA-KITTI and the real-world DENSE benchmark, as shown in Fig. 12. All methods follow the same evaluation protocol as in the main experiments. DDCD is shown as the filled polygon, with its raw RMSE reported at each vertex.

*Weather and severity.* DDCD achieves the most consistent robustness across adverse conditions. On AURORA-KITTI, it outperforms the completion baselines on every rain, snow, and fog category, with especially clear gains under severe corruption. For example, on Fog/Day, DDCD obtains 2.87 m RMSE, compared with 8.29–12.5 m for the completion baselines. The only adverse-weather case where this margin shrinks is heavy fog, where DDCD becomes comparable to the monocular UniK3D (4.32 vs. 4.29 m), suggesting that near-zero visibility substantially weakens the benefit of sparse LiDAR. A similar trend is observed on DENSE: DDCD is best on most weather×time and severity axes, and remains particularly strong under fog, e.g., Fog/Day 2.46 m vs. UniK3D 3.02 m. Across both datasets, fog severity forms the clearest difficulty gradient, while rain severity is a weaker driver of error.

*Time of day.* DDCD is also more stable between day and night than most baselines. On AURORA-KITTI, the night-time increase remains small across adverse weather, indicating that the model does not rely solely on clean RGB appearance cues. On DENSE, DDCD remains nearly symmetric on clear scenes and degrades only moderately under fog. The main exception is Rain/Night, where the monocular UniK3D performs better. This suggests that, when rain-induced LiDAR corruption is sparse and geometrically plausible, LiDAR conditioning can be less reliable than a strong image-only metric prior.

*Failure cases.* The cases where DDCD does not lead are interpretable. On AURORA-KITTI, it is slightly worse only in clean scenes, where there is no weather corruption to suppress, and it is comparable to UniK3D under heavy fog. On DENSE, its main weakness is rain, especially at night. Overall, DDCD trades a small amount of clean-weather accuracy for substantially stronger robustness under fog, snow, and most adverse-weather conditions, without collapsing on any specific category.

## 5.2. Boosting Robustness via Synthetic Data

To validate the effectiveness of AURORA-KITTI for robustness enhancement, we retrain representative supervised depth completion baselines on the synthetic training split and evaluate them on both the AURORA-KITTI test set and the real-world DENSE benchmark. For fair comparison, inputs from DENSE are pre-processed to the KITTI resolution to ensure a consistent evaluation protocol. Table 7 summarises the results before and after retraining. Overall, synthetic adverse-weather training yields substantial gains on AURORA-KITTI and consistent improvements on DENSE, demonstrating both strong in-domain adaptation and promising cross-domain generalisation.

On the AURORA-KITTI test set, retraining markedly improves both Sparse2Dense and BP-Net across all evaluation metrics. For Sparse2Dense, RMSE decreases from 4660.44 mm to 2497.56 mm, while BP-Net improves from 4945.43 mm to 2103.33 mm. Similar trends are observed for MAE, iRMSE, and iMAE, confirming that exposure to synthetic weather degradation substantially improves robustness under the proposed DCD setting. The gains also transfer to the real-world DENSE dataset. After retraining, Sparse2Dense reduces RMSE from 5416.31 mm to 4973.36 mm, while BP-Net improves from 4946.62 mm to 4252.91 mm, with corresponding improvements in the remaining metrics. These results indicate that training on AURORA-KITTI not only improves in-domain performance but also enhances robustness under real adverse weather.

Fig. 11 provides qualitative comparisons between pretrained models (top row) and models retrained on AURORA-KITTI (bottom row). On both DENSE and AURORA-KITTI examples, retrained models exhibit clearer object boundaries, improved scale alignment, and reduced noise artefacts. In particular, distant objects and foreground structures become more distinguishable after retraining, and large-scale depth compression artefacts observed in pretrained models are significantly mitigated. These visual improvements corroborate the quantitative gains reported in Table 7. Overall, the results demonstrate that our synthetic AURORA-KITTI data effectively enhances robustness for supervised depth completion models. Sparse2Dense shows strong data sensitivity and benefits substantially from in-domain retraining, while BP-Net exhibits better cross-domain stability and achieves more consistent improvements on the real-world DENSE benchmark. These complementary behaviors highlight the value of synthetic adverse-weather data for improving both in-domain accuracy and cross-domain generalisation in depth completion and denoising under challenging conditions.

For clarity, Table 8 places the existing in-domain results side by side; no additional models were trained for this comparison. Retraining the two original backbones measures the gain from AURORA-KITTI exposure. Holding the BP-Net student backbone and underlying training split fixed, DDCD further reduces AURORA-KITTI RMSE from 2103.33 to 1801.93 mm. This additional gain belongs to the complete DDCD training pipeline, including its privileged paired clean-RGB teacher supervision, and should not be interpreted as an architecture-only effect under identical supervision.

**Table 8**
In-domain RMSE comparison of existing AURORA-KITTI-trained models on the AURORA-KITTI and DENSE test sets. Sparse2Dense and BP-Net are retrained without teacher distillation; DDCD uses a BP-Net student and paired clean-RGB teacher supervision only during training. The BP-Net and DDCD rows share the student backbone and underlying training split, but differ in their supervision.

| Method | Clean | AURORA-KITTI↓ | DENSE↓ |
|---|---|---|---|
| Sparse2Dense (retrained) | No | 2497.56 | 4973.36 |
| BP-Net (retrained) | No | 2103.33 | 4252.91 |
| DDCD (BP-Net) | Yes | 1801.93 | 2761.97 |

**Table 9**
Teacher ablation using Depth Anything V2 and Depth Anything V3 as the depth foundation model for clean-prior distillation. The top two rows report AURORA-KITTI results and the bottom two rows report DENSE results.

| Model | RMSE↓ | MAE↓ | iRMSE↓ | iMAE↓ |
|---|---|---|---|---|
| DepthAnything2 | 1801.93 | 570.36 | 4.25 | 1.96 |
| DepthAnything3 | **1711.78** | **524.15** | **3.99** | **1.90** |
| DepthAnything2 | **2761.97** | **1569.14** | **34.82** | **13.24** |
| DepthAnything3 | 3950.11 | 2604.12 | 41.72 | 23.54 |

**Table 10**
Component-wise ablation on AURORA-KITTI. "Data" denotes retraining on synthetic adverse-weather data; "Distil." denotes the structural-prior loss ($\mathcal{L}_{ssi} + \mathcal{L}_{grad}$); and "Norm." denotes the GT-aligned teacher normalisation in Eq. (6). A uses Data only; B adds a noisy-RGB teacher with Norm.; C uses a clean-RGB teacher without Norm. The clean-RGB prior is training-only.

| Conf. | Teacher | Data | Distil. | Norm. | RMSE↓ | MAE↓ | iRMSE↓ | iMAE↓ |
|---|---|---|---|---|---|---|---|---|
| Base | – | × | × | × | 4945.43 | 1818.58 | 12.32 | 5.73 |
| + A | – | ✓ | × | × | 2103.33 | 598.72 | 4.51 | 2.46 |
| + B | Noisy | ✓ | ✓ | ✓ | 1956.83 | 580.46 | 4.49 | 2.17 |
| + C | Clean | ✓ | ✓ | × | 1842.84 | 574.74 | 4.37 | **1.64** |
| **DDCD** | Clean | ✓ | ✓ | ✓ | **1801.93** | **570.36** | **4.25** | 1.96 |

## 5.3. Ablation study

We conduct two ablation studies to analyse the impact of (1) different depth foundation models used for clean-pair distillation and (2) each component in our training strategy, including synthetic data retraining and depth-prior distillation.

Table 9 compares using Depth Anything V2 and Depth Anything V3 as the depth foundation model to generate clean depth priors for distillation. The top two rows report results on AURORA-KITTI, while the bottom two rows correspond to DENSE. On the AURORA-KITTI test set, Depth Anything V3 slightly outperforms V2. Specifically, RMSE decreases from 1801.93 mm (v2) to 1711.78 mm (v3). Similar trends are observed for MAE and inverse-depth metrics, suggesting that V3 provides marginally better in-domain priors for synthetic adverse-weather scenes. However, on the real-world DENSE dataset, the trend reverses. Models distilled using Depth Anything V2 achieve lower RMSE compared to those using v3. This observation is consistent with the direct benchmarking results reported in Table 4, where Depth Anything V3 exhibits higher errors on DENSE (7778.43 mm). These observations suggest that although V3 is stronger in metric depth estimation, its predictions suffer from a larger domain gap under unseen severe adverse-weather conditions. Overall, the two foundation models show comparable in-domain performance, but V2 provides better robustness when transferring to real-world adverse conditions.

Table 10 provides a component-wise ablation to isolate the contribution of data, distillation, and teacher normalisation. The largest improvement comes from adverse-weather retraining: compared with the zero-shot BP-Net baseline, adding AURORA-KITTI training data reduces RMSE from 4945.43 mm to 2103.33 mm and MAE from 1818.58 mm to 598.72 mm. On top of this, structural-prior distillation with a noisy-RGB teacher further improves RMSE to 1956.83 mm, indicating that the teacher prior still provides useful geometric guidance under degraded inputs. Replacing the noisy-RGB teacher with the paired clean-RGB teacher further improves the results, while the clean image is used only as a training-time supervision source and is never used at inference. Finally, GT-aligned teacher normalisation yields the best RMSE, MAE, and iRMSE, improving RMSE from 1842.84 mm to 1801.93 mm and iRMSE from 4.37 to 4.25 compared with the unnormalised clean-teacher variant. Although the unnormalised variant obtains slightly lower iMAE, DDCD achieves the best overall dense-depth accuracy, showing that the final gain comes from the combination of adverse-weather data, structural distillation, and teacher-prior normalisation.

## 5.4. Discussion

Our experiments on AURORA-KITTI and the real-world DENSE benchmark offer several insights into robust depth perception under adverse weather.

*Foundation priors are strong but not uniformly robust.* Depth foundation models deliver compelling structural reconstructions on AURORA-KITTI, indicating that large-scale pretraining provides transferable geometric priors. However, cross-domain results on DENSE reveal that stronger in-domain performance does not necessarily translate to robustness under severe scattering and low visibility. In particular, priors generated by different foundation teachers lead to notably different transfer behavior (e.g., Depth Anything V2 vs. V3), suggesting that robustness is governed not only by model capacity but also by how the learned priors respond to distribution shift in extreme weather.

*Synthetic adverse-weather exposure is the primary robustness driver.* Retraining supervised completion models on AURORA-KITTI yields the dominant gains, highlighting that robustness is largely achieved by aligning the training distribution with realistic weather-induced corruptions. This conclusion is supported by a two-factor comparison. When the architecture is fixed and only the training data is changed, AURORA-KITTI retraining yields a large gain: BP-Net improves from 4945.43 to 2103.33 mm RMSE, and Sparse2Dense from 4660.44 to 2497.56 mm (Table 7). The zero-shot results in Table 3 describe performance under different training protocols and cannot isolate the effects of architecture and training data. Table 10 further confirms this trend within our own pipeline: weather-aware retraining provides the dominant improvement, while clean-prior distillation adds smaller but complementary gains. The in-domain comparison in Table 8 shows a further 301.40 mm (14.3 percent) reduction in AURORA-KITTI RMSE from the retrained BP-Net to DDCD, which shares its student backbone and underlying training split but additionally uses clean-RGB teacher supervision. This supports an incremental benefit of the DDCD training pipeline beyond data exposure, while the larger zero-shot-to-in-domain gaps cannot be credited solely to the method. These results indicate that architecture matters, but physics-consistent weather-aware data is the primary driver of robustness in our setting. Depth-prior distillation further improves performance, but mainly as a complementary regulariser: it stabilises global geometry and metric scale, and is particularly beneficial for near-range accuracy (inverse-depth metrics) where weather noise most strongly impacts downstream robotics tasks.

*Architecture and efficiency matter for deployment.* We observe complementary behaviors across model families: lightweight U-Net-style fusion models (e.g., Sparse2Dense) are highly data-sensitive and benefit substantially from in-domain retraining, whereas BP-Net exhibits more stable cross-domain improvements on DENSE, indicating stronger transferability under unseen weather. Meanwhile, diffusion-based approaches (e.g., Marigold-DC) can achieve strong absolute errors but incur prohibitive latency due to iterative refinement, limiting real-time applicability despite their accuracy.

*DCD requires explicit denoising, not just completion.* Across methods, treating weather-corrupted sparse LiDAR or RGB as standard completion inputs often propagates artefacts into dense predictions. By explicitly targeting a clean, dense depth output and injecting clean structural priors during training, DDCD consistently improves both in-domain performance and real-world transfer, validating the necessity of jointly modelling completion and denoising under adverse weather.

# 6. Conclusion

We introduce **AURORA-KITTI**, a large-scale synthetic benchmark for any-weather depth completion and denoising (DCD) with weather-consistent RGB–LiDAR degradation and paired clean references and metric depth supervision. To support the proposed DCD formulation, we further introduce LiDAR-aware denoising metrics, including ERR, DCE, and CPP, which complement standard depth-completion metrics by explicitly measuring corrupted-return correction, dropout-region recovery, and reliable-return preservation.

Through systematic benchmarking on AURORA-KITTI and evaluation on the real-world **DENSE** pixel-accurate benchmark, we show that existing depth completion pipelines and depth foundation models remain vulnerable under severe adverse weather, and that weather-aware, physically consistent training data provides the dominant robustness gains in our controlled evaluation. We further provide **DDCD** as a strong and efficient in-domain reference baseline, a distillation-based framework that transfers clean structural priors from frozen depth foundation teachers to stabilise metric scale and suppress weather-induced artefacts. Its paired clean-RGB teacher is used only during training and provides privileged supervision unavailable to zero-shot baselines; at inference, DDCD uses only degraded RGB and sparse LiDAR.

DDCD provides complementary improvements in structural consistency and metric alignment beyond weather-aware retraining, while maintaining practical inference efficiency. The benchmark tables compare models trained under different protocols and thus do not isolate an architectural advantage. The same-data BP-Net comparison and component ablations support a further, smaller gain from DDCD beyond weather-aware retraining. By design, AURORA-KITTI serves as a controllable training and benchmarking proxy with paired clean references and physically interpretable weather controls; it is intended to complement, rather than replace, real adverse-weather capture. We hope AURORA-KITTI, the LiDAR-aware DCD metrics, and the proposed DCD protocol facilitate future research on robust, weather-aware depth perception and cross-domain generalisation.

## References

[1] J. Guo, J. Ma, F. Sun, Z. Gao, Á. F. García-Fernández, H.-N. Liang, X. Zhu, W. Ding, Cd-udepth: Complementary dual-source information fusion for underwater monocular depth estimation, Information Fusion 118 (2025) 102961.

[2] Y. He, P. Yin, F. R. Yu, X. Zeng, Z. Liu, Light-weight monocular depth estimation via transformer-fusion for monocular visual slam, Information Fusion (2025) 103591.

[3] D. Fernandes, A. Silva, R. Névoa, C. Simões, D. Gonzalez, M. Guevara, P. Novais, J. Monteiro, P. Melo-Pinto, Point-cloud based 3d object detection and classification methods for self-driving applications: A survey and taxonomy, Information Fusion 68 (2021) 161–191.

[4] K. Wu, J. Lin, J. Miao, Z. Li, X. Zhang, G. Xing, Y. Fan, J. Luo, H. Zhao, Y. Liu, et al., Difnet: Dual-information fusion network for depth completion, Information Fusion 125 (2026) 103424.

[5] F. Ma, G. V. Cavalheiro, S. Karaman, Self-supervised sparse-to-dense: Self-supervised depth completion from lidar and monocular camera, in: 2019 International Conference on Robotics and Automation (ICRA), 2019, pp. 3288–3295. doi:10.1109/ICRA.2019.8793637.

[6] Y. Wang, G. Zhang, S. Wang, B. Li, Q. Liu, L. Hui, Y. Dai, Improving depth completion via depth feature upsampling, in: Proceedings of the IEEE/CVF Conference on Computer Vision and Pattern Recognition, 2024, pp. 21104–21113.

[7] J. Tang, F.-P. Tian, B. An, J. Li, P. Tan, Bilateral propagation network for depth completion, in: Proceedings of the IEEE/CVF Conference on Computer Vision and Pattern Recognition (CVPR), 2024, pp. 9763–9772.

[8] J. Uhrig, N. Schneider, L. Schneider, U. Franke, T. Brox, A. Geiger, Sparsity invariant cnns, in: International Conference on 3D Vision (3DV), 2017.

[9] T. Gruber, M. Bijelic, F. Heide, W. Ritter, K. Dietmayer, Pixel-accurate depth evaluation in realistic driving scenarios, in: International Conference on 3D Vision (3DV), 2019.

[10] J. Zhang, R. Liu, H. Shi, K. Yang, S. Reiß, K. Peng, H. Fu, K. Wang, R. Stiefelhagen, Delivering arbitrary-modal semantic segmentation, in: Proceedings of the IEEE/CVF Conference on Computer Vision and Pattern Recognition, 2023, pp. 1136–1147.

[11] J. Wang, C. Lin, L. Nie, S. Huang, Y. Zhao, X. Pan, R. Ai, Weatherdepth: Curriculum contrastive learning for self-supervised depth estimation under adverse weather conditions, in: 2024 IEEE International Conference on Robotics and Automation (ICRA), IEEE, 2024, pp. 4976–4982.

[12] X. Yan, C. Zheng, Y. Xue, Z. Li, S. Cui, D. Dai, Benchmarking the robustness of lidar semantic segmentation models, International Journal of Computer Vision 132 (7) (2024) 2674–2697.

[13] T. Brödermann, D. Bruggemann, C. Sakaridis, K. Ta, O. Liagouris, J. Corkill, L. Van Gool, Muses: The multi-sensor semantic perception dataset for driving under uncertainty, in: European Conference on Computer Vision, Springer, 2024, pp. 21–38.

[14] Y. Wang, H. Zhao, D. Gummadi, M. Dianati, K. Debattista, V. Donzella, Robustness of panoptic segmentation for degraded automotive cameras data, IEEE Transactions on Automation Science and Engineering 22 (2025) 21020–21032. doi:10.1109/TASE.2025.3605753.

[15] M. Hu, S. Wang, B. Li, S. Ning, L. Fan, X. Gong, Penet: Towards precise and efficient image guided depth completion, in: 2021 IEEE International Conference on Robotics and Automation (ICRA), IEEE, 2021, pp. 13656–13662.

[16] X. Cheng, P. Wang, C. Guan, R. Yang, Cspn++: Learning context and resource aware convolutional spatial propagation networks for depth completion, in: Proceedings of the AAAI conference on artificial intelligence, Vol. 34, 2020, pp. 10615–10622.

[17] J. Park, K. Joo, Z. Hu, C.-K. Liu, I. So Kweon, Non-local spatial propagation network for depth completion, in: European conference on computer vision, Springer, 2020, pp. 120–136.

[18] Y. Zhang, X. Guo, M. Poggi, Z. Zhu, G. Huang, S. Mattoccia, Completionformer: Depth completion with convolutions and vision transformers, in: Proceedings of the IEEE/CVF Conference on Computer Vision and Pattern Recognition (CVPR), 2023, pp. 18527–18536.

[19] Z. Yan, Y. Lin, K. Wang, Y. Zheng, Y. Wang, Z. Zhang, J. Li, J. Yang, Tri-perspective view decomposition for geometry-aware depth completion, in: Proceedings of the IEEE/CVF Conference on Computer Vision and Pattern Recognition, 2024, pp. 4874–4884.

[20] Y. Zuo, W. Yang, Z. Ma, J. Deng, Omni-dc: Highly robust depth completion with multiresolution depth integration, in: Proceedings of the IEEE/CVF International Conference on Computer Vision, 2025, pp. 9287–9297.

[21] R. Ranftl, K. Lasinger, D. Hafner, K. Schindler, V. Koltun, Towards robust monocular depth estimation: Mixing datasets for zero-shot cross-dataset transfer, IEEE transactions on pattern analysis and machine intelligence 44 (3) (2020) 1623–1637.

[22] L. Piccinelli, Y.-H. Yang, C. Sakaridis, M. Segu, S. Li, L. Van Gool, F. Yu, Unidepth: Universal monocular metric depth estimation (2024).

[23] L. Yang, B. Kang, Z. Huang, X. Xu, J. Feng, H. Zhao, Depth anything: Unleashing the power of large-scale unlabeled data, in: Proceedings of the IEEE/CVF conference on computer vision and pattern recognition, 2024, pp. 10371–10381.

[24] L. Yang, B. Kang, Z. Huang, Z. Zhao, X. Xu, J. Feng, H. Zhao, Depth anything v2, arXiv:2406.09414 (2024).

[25] H. Lin, S. Chen, J. H. Liew, D. Y. Chen, Z. Li, G. Shi, J. Feng, B. Kang, Depth anything 3: Recovering the visual space from any views, arXiv preprint arXiv:2511.10647 (2025).

[26] Y. Liang, Y. Hu, W. Shao, Y. Fu, Distilling monocular foundation model for fine-grained depth completion, in: Proceedings of the Computer Vision and Pattern Recognition Conference, 2025, pp. 22254–22265.

[27] T. Liang, H. Xie, K. Yu, Z. Xia, Z. Lin, Y. Wang, T. Tang, B. Wang, Z. Tang, Bevfusion: A simple and robust lidar-camera fusion framework, Advances in neural information processing systems 35 (2022) 10421–10434.

[28] Y. Cui, R. Chen, W. Chu, L. Chen, D. Tian, Y. Li, D. Cao, Deep learning for image and point cloud fusion in autonomous driving: A review, IEEE Transactions on Intelligent Transportation Systems 23 (2) (2021) 722–739.

[29] R. Liu, G. Zhang, J. Wang, S. Zhao, Cross-modal 360 depth completion and reconstruction for large-scale indoor environment, IEEE Transactions on Intelligent Transportation Systems 23 (12) (2022) 25180–25190.

[30] R. Liu, Y. Qin, Y. Pan, Q. Li, B. Sun, J. Zhang, Panorama-lidar fusion for dense omnidirectional depth completion in 3d street scene, IEEE Transactions on Consumer Electronics 70 (2) (2024) 4756–4766.

[31] M. Wang, Y. Qin, R. Li, Z. Liu, Z. Tang, K. Li, Awdepth: Monocular depth estimation for adverse weather via masked encoding, IEEE Transactions on Industrial Informatics 20 (9) (2024) 10873–10882.

[32] Z. Yan, Z. Wang, K. Wang, J. Li, J. Yang, Completion as enhancement: A degradation-aware selective image guided network for depth completion, in: Proceedings of the Computer Vision and Pattern Recognition Conference, 2025, pp. 26943–26953.

[33] M. Viola, K. Qu, N. Metzger, B. Ke, A. Becker, K. Schindler, A. Obukhov, Marigold-dc: Zero-shot monocular depth completion with guided diffusion, in: Proceedings of the IEEE/CVF International Conference on Computer Vision, 2025, pp. 5359–5370.

[34] K. Saunders, G. Vogiatzis, L. J. Manso, Self-supervised monocular depth estimation: Let's talk about the weather, in: Proceedings of the IEEE/CVF International Conference on Computer Vision, 2023, pp. 8907–8917.

[35] M. Tremblay, S. S. Halder, R. De Charette, J.-F. Lalonde, Rain rendering for evaluating and improving robustness to bad weather, International Journal of Computer Vision 129 (2021) 341–360.

[36] V. Kilic, D. Hegde, A. B. Cooper, V. M. Patel, M. Foster, Lidar light scattering augmentation (lisa): Physics-based simulation of adverse weather conditions for 3d object detection, in: ICASSP 2025-2025 IEEE International Conference on Acoustics, Speech and Signal Processing (ICASSP), IEEE, 2025, pp. 1–5.

[37] M. Hahner, C. Sakaridis, M. Bijelic, F. Heide, F. Yu, D. Dai, L. Van Gool, Lidar snowfall simulation for robust 3d object detection, in: Proceedings of the IEEE/CVF conference on computer vision and pattern recognition, 2022, pp. 16364–16374.

[38] M. Hahner, C. Sakaridis, D. Dai, L. Van Gool, Fog simulation on real lidar point clouds for 3d object detection in adverse weather, in: Proceedings of the IEEE/CVF international conference on computer vision, 2021, pp. 15283–15292.

[39] C. Sakaridis, D. Dai, L. Van Gool, Semantic foggy scene understanding with synthetic data, International Journal of Computer Vision 126 (9) (2018) 973–992.

[40] Y. Wang, B. Li, Z. Wei, A. Rahman, D. Gummadi, H. Zhao, V. Donzella, A survey and new perspective of sensing in the dark for intelligent transportation systems, IEEE Transactions on Intelligent Transportation Systems 26 (11) (2025) 18283–18303. doi:10.1109/TITS.2025.3600557.

[41] J. Ku, A. Harakeh, S. L. Waslander, In defense of classical image processing: Fast depth completion on the cpu, in: 2018 15th Conference on Computer and Robot Vision (CRV), 2018, pp. 16–22. doi:10.1109/CRV.2018.00013.

[42] L. Piccinelli, C. Sakaridis, M. Segu, Y.-H. Yang, S. Li, W. Abbeloos, L. Van Gool, UniK3D: Universal camera monocular 3d estimation, in: IEEE/CVF Conference on Computer Vision and Pattern Recognition (CVPR), 2025.